\documentclass[ijoc,sglanonrev]{informs4}
\usepackage{eqndefns-left} 
\RequirePackage{tgtermes}
\RequirePackage{newtxtext}
\RequirePackage{newtxmath}
\RequirePackage{bm}
\RequirePackage{endnotes}

 \OneAndAHalfSpacedXII 
 
\usepackage{natbib}
\bibpunct[, ]{(}{)}{,}{a}{}{,}%
\def\bibfont{\small}%

 \usepackage{array}
 \usepackage{booktabs}
 \usepackage{graphicx}
 \usepackage{subfigure}
 \usepackage{float}
 \usepackage{threeparttable}
 \usepackage{verbatim}

 \usepackage{amsfonts}
 \usepackage{algorithm}
 \usepackage{algorithmic}
 \usepackage{indentfirst}
 \usepackage{float}
 
 \usepackage{url}
 \usepackage{caption}
 \usepackage{bbding} 

 \usepackage{colortbl} 
 \usepackage{xcolor}
 \usepackage{array}
 \usepackage{color}
 
 \usepackage{bbm}
 \usepackage[english]{babel}
 \usepackage[letterpaper,top=1in,bottom=1in,left=1in,right=1in]{geometry}
 \usepackage[colorlinks=true, allcolors=blue]{hyperref}
 \usepackage{amsmath}
 \usepackage{amssymb}
 
 \usepackage{xspace}
 \usepackage{csquotes}
 \usepackage{makecell}
 \usepackage{multirow}
 
\usepackage{pifont}    

\EquationsNumberedThrough    

\TheoremsNumberedThrough     
\ECRepeatTheorems  %

\begin{document}




\TITLE{Striking the Perfect Balance: Preserving Privacy While Boosting Utility in Collaborative Medical Prediction Platforms}

%


\ARTICLEAUTHORS{%
	\AUTHOR{
    Shao-Bo Lin, Xiaotong Liu\thanks{corresponding author: ariesoomoon@gmail.com}, Yao Wang }
	\AFF{ Center for Intelligent Decision-Making and Machine Learning, School of Management, Xi'an Jiaotong University, Xi'an, China}
}

\ABSTRACT{%
Online collaborative medical prediction platforms offer convenience and real-time feedback by leveraging massive electronic health records.  However, growing concerns about privacy and low prediction quality can deter patient participation and doctor cooperation.   In this paper, we first clarify the privacy attacks, namely attribute attacks targeting patients and model extraction attacks targeting doctors, and specify the corresponding privacy principles. We then propose a privacy-preserving mechanism and integrate it into a novel one-shot distributed learning framework, aiming to simultaneously meet both privacy requirements and prediction performance objectives. Within the framework of statistical learning theory, we theoretically demonstrate that the proposed distributed learning framework can achieve the optimal prediction performance under specific privacy requirements. We further validate the developed privacy-preserving collaborative medical prediction platform through both toy simulations and real-world data experiments.
}%





\maketitle


\section{Introduction}\label{sec:Intro}
The advent of the Internet era has brought about profound changes, shifting management models, business practices, and even people's lifestyles from offline to online modes. With the help of large electronic health records (EHRs),
online medical prediction platforms (OMPPs) such as iCliniq, Zocdoc, and DOCTO  offer significant convenience and flexibility by breaking the geographical barriers of traditional medical consultations \citep{yan2014feeling}. Despite their increasingly important role in people’s daily lives, OMPPs also introduce several challenges, including the dissemination of inaccurate information, rising health-related anxiety, declining patient satisfaction, low-quality predictions, and, most notably, growing concerns over privacy issues \citep{keshta2021security}. Actually, millions of pregnant and postpartum mothers was leaked, disrupting the lives of newborn families\footnote{https://www.yifahui.com/2432.html}, and personal information  of 1.41 million U.S. doctors from the FAD platform  was sold on a hacker forum\footnote{ https://hackread.com/
personal-data-us-doctors-sold-hacker-forum/}. As highlighted by \cite{antheunis2013patients}, these serious privacy breaches are considered the primary obstacle to the development of OMPPs.

\begin{figure}[H]
	\centering
	\caption{Different Online Medical Prediction Modes.}
	\vspace{0.1in}
	\setlength{\subfigcapskip}{-0.5em}
	\subfigure[One-to-one mode]{\includegraphics[scale=0.2]{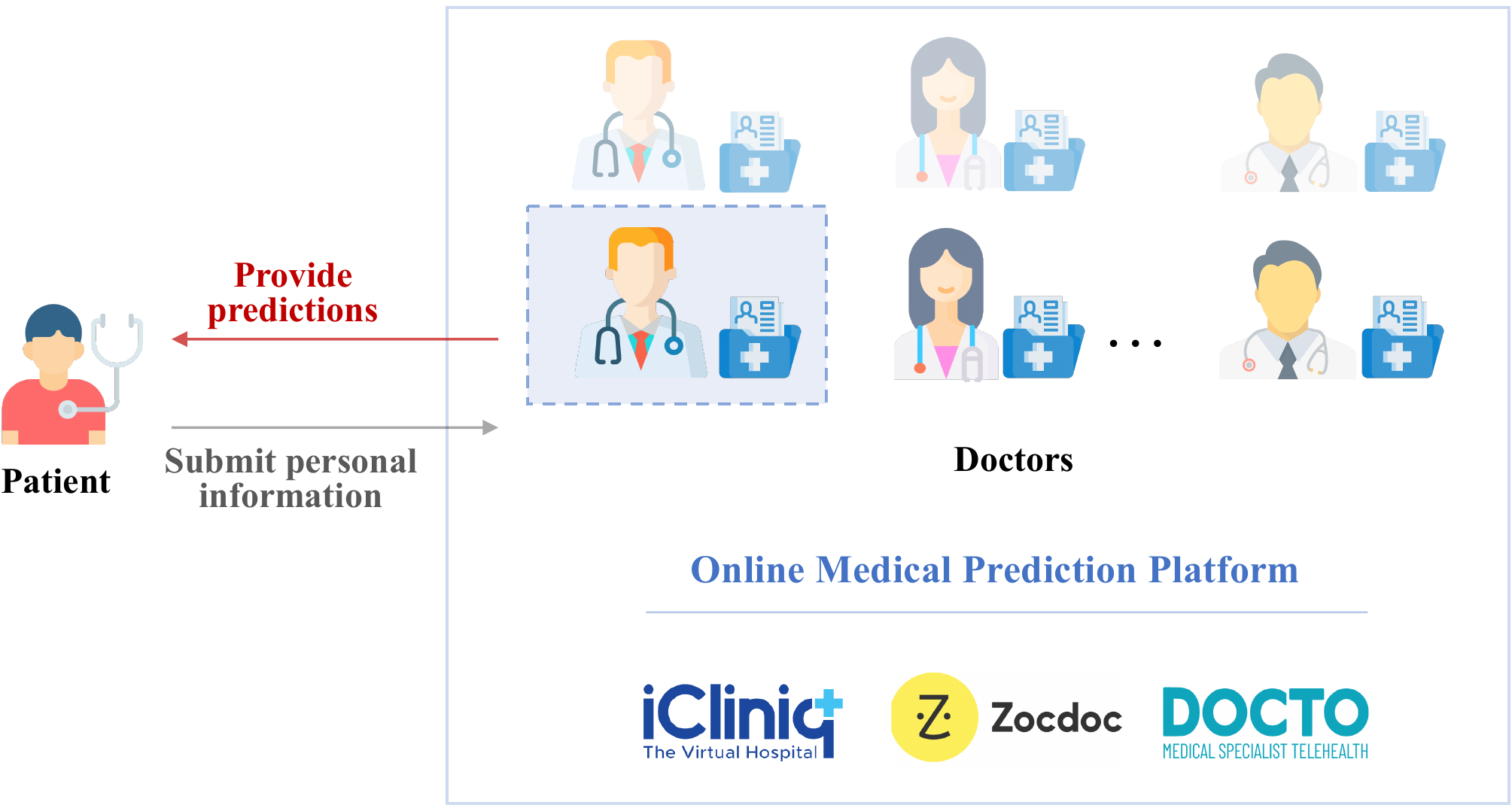}
		\label{subfig: onetoone}} 
	\setlength{\subfigcapskip}{-0.5em}
	\subfigure[One-to-many mode]{\includegraphics[scale=0.2]{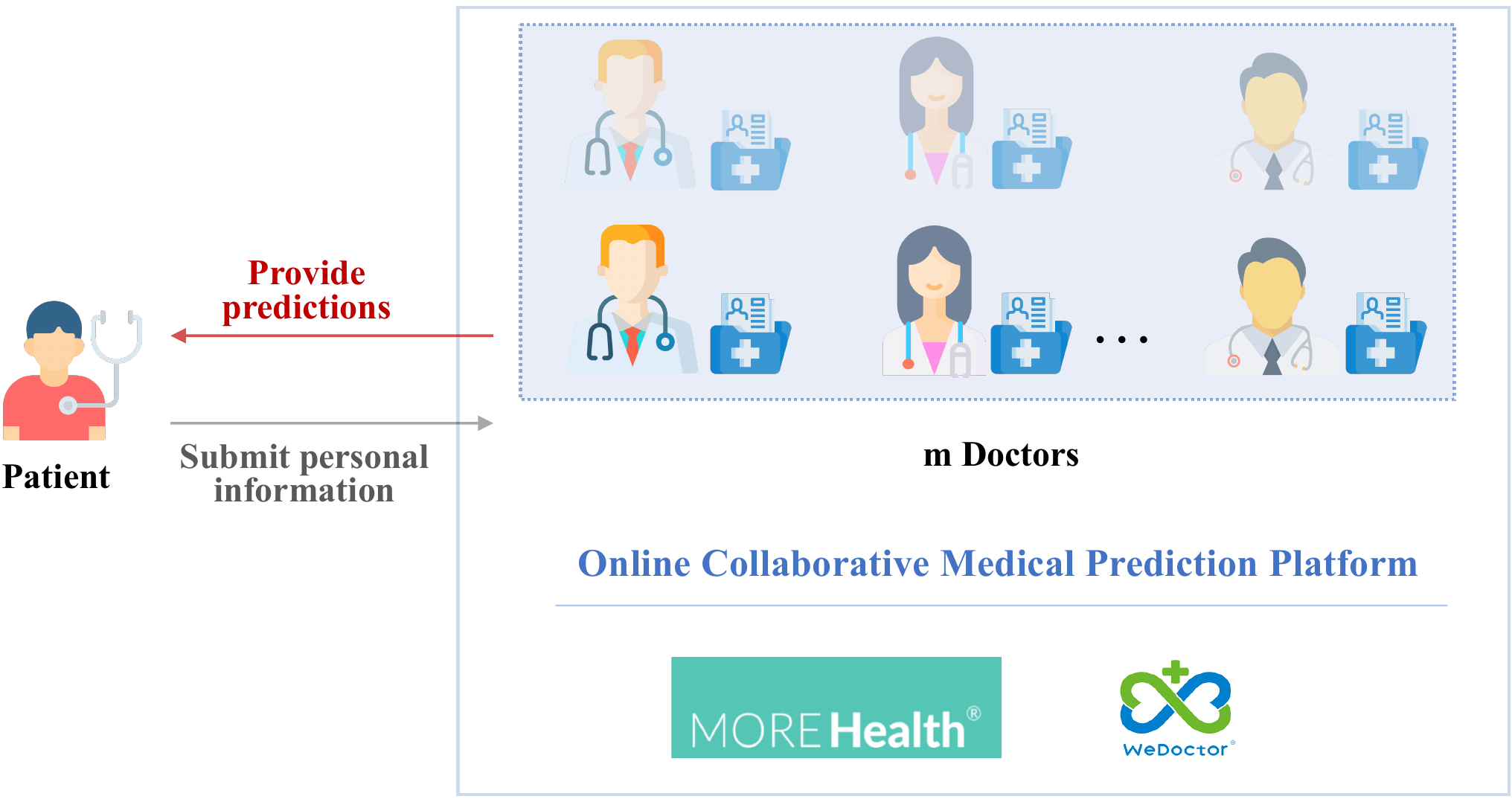}
		\label{subfig: onetomany}} 
	
	\label{fig:onetoone_many}
\end{figure}
\vspace{-0.5cm}

As medical prediction places significant emphasis on accuracy due to the potential adverse consequences of errors \citep{liu2022machine, ray2023lstm}, the traditional one-to-one mode, as illustrated in Figure \ref{fig:onetoone_many}(a), often fails to deliver high-quality predictions \citep{huang2019patient}, primarily due to the employment of inexperienced doctors.
Online collaborative medical prediction platforms (CMPPs), such as MORE Health and WeDoctor, have been developed to enhance prediction quality by engaging multiple doctors to serve a single patient, as shown in Figure \ref{fig:onetoone_many}(b), using federated learning or distributed learning techniques \citep{deist2020distributed, zhou2020differentially,liu2022enabling}.
However, with multiple doctors accessing patient information, the privacy issue in CMPPs become more serious in the sense that it is difficult to judge which doctors are unreliable. Additionally, doctors may hesitate to participate due to concerns that their decision-making processes (or models), repeatedly engaged across diagnostic tasks, may be exposed to model extraction attacks \citep{tramer2016stealing}. Under this circumstance, it is highly desired to develop practical privacy-preserving mechanisms to equip CMPP to tackle the   privacy issues  without sacrificing prediction performance.

Several efforts have been made to address the privacy issues in CMPP, including a privacy-preserving distributed clinical decision support system \citep{mathew2011privacy} to conceal patients’ personal information,  a homomorphic encryption and secure multi-party healthcare system \citep{zhang2022homomorphic} to prevent adversaries from stealing doctors’ models, and the PriMIA framework \citep{kaissis2021end} that integrates differentially private federated model  with encrypted aggregation   to safeguard doctors' models from disclosure. These pioneering studies provide valuable guidance for developing privacy-preserving CMPP (PPCMPP), significantly advancing the practical development and application of CMPP.

However, unilateral privacy preservation for doctors or patients alone cannot meet the dual privacy requirements   for both doctors and patients in CMPP, resulting in an critical gap between existing approaches and  the practical demands of PPCMPP.  
 Moreover, directly combining these unilateral privacy-preserving methods, such as \citep{mathew2011privacy} and \citep{zhang2022homomorphic}, is infeasible for achieving PPCMPP with dual privacy requirements, as they are designed for different algorithms (e.g., decision trees in \citep{mathew2011privacy} and deep learning in \citep{zhang2022homomorphic}). 
Even setting aside these technical mismatches, such straightforward combinations fail to quantify the relationship between privacy preservation and prediction accuracy — a limitation that is unacceptable in medical prediction tasks, where extremely high accuracy is required. Our goal is to design a novel PPCMPP that simultaneously fulfills the dual privacy requirements of both doctors and patients without compromising prediction accuracy.

\subsection{Road-map and Our Approach}
As the approaches in the existing literature \citep{mathew2011privacy,dayan2021federated,zhang2022homomorphic,kaissis2021end} focus  solely on the privacy strategies  without considering the different privacy attacks targeting patients and doctors, we begin by qualifying the privacy attacks and the corresponding privacy principles that evaluate how well the privacy is protected. Since patients submit their personal information to the CMPP for query, attackers (potentially unethical doctors) are capable of inferring  their identities by linking attributes like race, age, and weight with publicly available data, corresponding to the well-known attribute attack \citep{machanavajjhala2007diversity,li2006t}. 
For doctors, CMPP can generate fake queries to which the targeted doctor provides responses; by collecting these input–output pairs, CMPP can reconstruct the doctor’s model and  replace the victim doctor with this constructed model. Model extraction attacks \citep{tramer2016stealing} then occur and the victim doctor is essentially forced to provide services but illegally  removed  from CMPP without  receiving any compensation. In summary, our focus is on defending against attribute attacks \citep{machanavajjhala2007diversity} on patients and model extraction attacks \citep{tramer2016stealing} on doctors.

Although numerous privacy principles have been proposed to measure the quality of privacy preservation against some specific attacks, with typical examples including $k$-anonymity \citep{sweeney2002k}, $l$-diversity \citep{machanavajjhala2007diversity}, and $t$-closeness \citep{li2006t} for linkage attacks, as well as differential privacy for probabilistic attacks \citep{dwork2008differential} and collusion attacks \citep{li2017differential}, appropriate principles for addressing attribute attacks and model extraction attacks in CMPP remain lacking. This gap is primarily due to the requirements of real-time preservation, evaluation, and feedback inherent to CMPP settings. In response, we propose a novel CO principle for attribute attacks and modify the existing RL principle \citep{Li2006} for model extraction attacks in CMPP.

Besides privacy considerations, utility, measured by prediction accuracy, is also crucial for evaluating the quality of CMPP and thus imposes strict restrictions on utility, excluding several widely adopted approaches, such as generalization \citep{sweeney2002k}, suppression \citep{samarati2002protecting}, microaggregation \citep{domingo2002practical}, and noising \citep{dwork2008differential}. 
The possibly contradictory high demands for privacy and utility correspond to a hard-to-solve optimization problem of maximizing prediction accuracy under given CO and RL levels.

\begin{figure}[h]
	\centering
	\caption{Roadmap of Our Approach.}
	\vspace{0.1in}
	\includegraphics[scale=0.6]{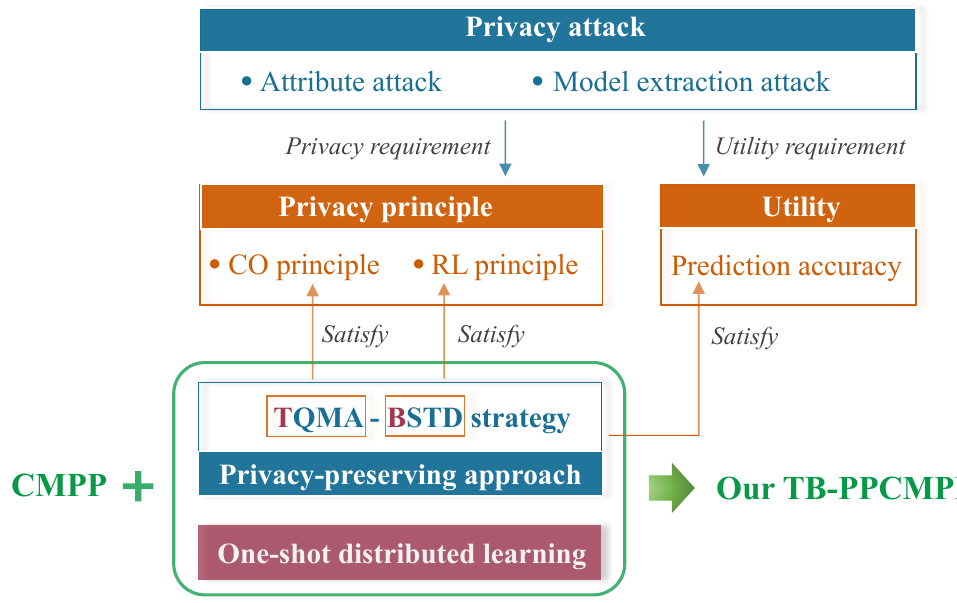}
	\label{Fig:roadmap}  
	\vspace{-0.3cm}
\end{figure}


Our approach is not to solve the mentioned optimization problem to pursue  a feasible privacy-preserving solution, but to present a novel one-shot  distributed learning framework with two interaction windows
to satisfy the specified utility and privacy requirements. Specifically, we first  introduce a tree-based binary subdivision strategy (called as TQMA) to defend against  attribute attacks, noting that tree-based localized regression methods are among the most widely used medical prediction algorithms by doctors \citep{nahar2018liver},
and then  combine  the ideas of bounded swapping from \citep{Li2011} and threshold decryption from \citep{lindell2005secure}
to develop a   bounded swapping and
threshold decryption (BSTD) mechanism to defend against model extraction attacks for doctors. With the help of privacy communications, these two approaches  are effectively 
 integrated within a  one-shot swapped distributed learning framework, resulting in a TQMA-BSTD–based PPCMPP (TB-PPCMPP) that successfully delivers predictions meeting both utility and privacy requirements. The roadmap of TB-PPCMPP is exhibited   in Figure \ref{Fig:roadmap}.

\subsection{Related Work}

CMPP based on federated learning and distributed learning schemes has been widely developed \citep{brisimi2018federated, huang2019patient, choudhury2020predicting, liu2022enabling, deist2020distributed} to improve prediction accuracy. However, these studies do not sufficiently address defenses against privacy attacks targeting both patients and doctors, posing challenges for the real-world deployment of these distributed learning schemes. \citep{li2020federatedC}.

On the patient side, privacy-preserving approaches such as generalization \citep{sweeney2002k}, suppression \citep{samarati2002protecting}, microaggregation \citep{domingo2002practical}, noising \citep{dwork2008differential}, and probability-based swapping \citep{li2009against} indeed have the potential to defend against attribute attacks. However, the implementation of these methods relies on access to the entire dataset, making them unsuitable for CMPP, where patients require real-time privacy preservation. Moreover, these approaches typically achieve privacy preservation at the expense of prediction performance, falling short of meeting the high accuracy demands of patients.
In the CMPP setting, most existing work focuses on exploring the privacy preservation of distributively stored patient datasets held by doctors \citep{burnap2012protecting,lai2023edge}, rather than directly addressing the privacy concerns of incoming patients — particularly under the constraint of maintaining prediction performance.
Among the most relevant studies, \cite{mathew2011privacy} proposed a privacy-preserving distributed clinical decision support system that constructs decision trees without exposing patient data. In this approach, both local data and queries are represented as graphs to capture the structural information of local records. Each site locally matches the query graph, summarizes the matched records, and sends only aggregated statistics to a central agent, which then builds the decision tree and returns it to the requester. In their setting,
the relationship between the graph and the final result is unclear, making the process non-transparent and difficult to trust for privacy- and accuracy-conscious patients.

On the doctor side, numerous approaches have been proposed for
privacy-preserving collaborative prediction to protect doctors' privacy \citep{dayan2021federated, kaissis2021end, zhang2022homomorphic, brisimi2018federated}.
We highlight several studies most relevant to our work.
\cite{dayan2021federated} employed a federated learning framework combined with differential privacy to protect distributively stored data, using information from multiple institutions to train a federated model for predicting the oxygen requirements of COVID-19 patients.
\cite{kaissis2021end} proposed the PriMIA framework, which integrates differentially private federated model training with encrypted aggregation of model updates to safeguard local data and models from disclosure;
\cite{zhang2022homomorphic} proposed a federated learning mechanism utilizing homomorphic encryption and secure multi-party computation for deep learning in healthcare systems, safeguarding private local medical data from adversaries.
However, in their setting, the prediction model is either publicly known or limited to a specified algorithm, which conflicts with the practical need of doctors to maintain algorithmic privacy and independently choose decision-making methods.  Furthermore, most preservation mechanisms rely on encryption technologies, which are typically inaccessible to individuals without cryptographic expertise and require resource-intensive computation \citep{hastings2023privacy}, making it challenging to provide the real-time feedback required by CMPP.

Research that simultaneously considers the privacy issues of both patients and doctors is scarce. Although \cite{mathew2011privacy} addresses privacy concerns on both sides, the unified privacy mechanism it employs is not specifically designed to defend against model extraction attacks targeting doctors, nor does it discuss prediction performance, thus failing to meet patients’ demands for high accuracy.

Compared with existing PPCMPP
\citep{mathew2011privacy,dayan2021federated,zhang2022homomorphic,kaissis2021end},
there are mainly three advantages of the proposed TB-PPCMPP. At first, TB-PPCMPP aims at developing privacy-preserving  for both patients and doctors which is out of the scope of existing work. Then, TB-PPCMPP is essentially attack-driven approach but the privacy attacks in existing work are unknown. Finally, TB-PPCMPP is theoretically and empirically proven to successfully defend against both attribute attacks and model extraction attacks without sacrificing prediction accuracy — a novel achievement, as prior literature consistently reports a trade-off between privacy and utility.

\subsection{Our Contributions}\label{Contributions}

We outline our contributions in three aspects: methodology development, theoretical novelty, and management implications.

\begin{itemize}
	\item \textit{Methodology development:}
We formulate the privacy issue in CMPP as an optimization problem aiming to achieve optimal prediction performance under specific privacy constraints.  
By analyzing privacy attacks and defining corresponding privacy principles, we embed the  optimization problem within a distributed learning framework and transform it into a solvable machine learning problem.  
Based on this, we develop the TQMA-BSTD-based distributed learning framework for privacy-preserving CMPP, which integrates a tree-based binary subdivision strategy (TQMA) to counter attribute attacks and a bounded swapping and threshold decryption mechanism (BSTD) to resist model extraction attacks. Such distributed learning framework ensures privacy preservation without sacrificing prediction performance.

	\item \textit{Theoretical novelty:} 
Our study unveils a groundbreaking theoretical insight: the conventional privacy–utility trade-off is not universally applicable.  
We rigorously prove that the proposed TB-PPCMPP achieves optimal prediction accuracy while simultaneously reducing the risk of attribute and model extraction attacks for both patients and doctors.  
Importantly, our findings do not contradict the conventional privacy–utility trade-off, which generally applies to a broader range of privacy attacks and notions of data utility, as our focus is specifically on selected privacy attacks and utility measured in terms of prediction performance.

	\item \textit{Management implication:} 
Our study offers important managerial implications for enhancing privacy in online collaborative prediction platforms.  
Specifically, platform managers should identify the privacy attacks most relevant to their context and select suitable privacy principles alongside the specific data utility concerns of their participants.  
Based on these principles, they can frame the privacy issue as an optimization problem, where the search for effective privacy-preserving approaches becomes a matter of solving this optimization problem.  
By focusing on specific privacy attacks, platforms have the potential to mitigate the traditionally strict privacy–utility trade-off and achieve the dual objective of privacy preservation and high prediction performance.

\end{itemize}

\subsection{Organization}\label{Organization}
The rest of this paper proceeds as follows. Section \ref{Sec:privacy issues} discusses the privacy issues in CMPP and introduces the corresponding privacy-preserving mechanisms: the TQMA mechanism for defending against attribute attacks and the BSTD mechanism for model extraction attacks. Section \ref{Sec:platform_now} presents the TQMA-BSTD-based distributed learning framework for CMPP, referred to as TB-PPCMPP. Section \ref{Sec.theory} investigates the theoretical properties of the proposed TB-PPCMPP. Section \ref{experiment} details the experiments conducted on both simulated and real-world datasets. Finally, Section \ref{Conclusions} concludes the paper. Additional experiments and theoretical proofs are provided in the Appendix.

\section{Privacy Issues of CMPP}\label{Sec:privacy issues}
This section discusses the privacy issues of CMPP concerning patients and doctors, respectively.

\subsection{Privacy Preservation against Attribute Attacks} 
Let $\tilde{x}=(x^{(1)},\dots,x^{(d^\diamond)})^T\in\mathbb I^{d^\diamond}:=[a,b]^{d^\diamond}$ for $a,b\in\mathbb R$  be the complete information of a patient. Generally speaking, there are three categories of  attributes  of $\tilde{x}$ \citep{sweeney2002k},  namely, identity attributes (IA),  $\tilde{x}_{IA,\tilde{d}}$, confidential attributes (CA), $\tilde{x}_{CA,\bar{d}}$,  and quasi-identifier attributes (QIA), $\tilde{x}_{QIA,d'}$, for 
$d',\bar{d},\tilde{d}\in\mathbb N$ and $d'+\bar{d}+\tilde{d}=d^\diamond$. For the privacy concerns, patients   only submit  an anonymized version of $\tilde{x}$, $x=(\tilde{x}_{QIA,d'},\tilde{x}_{CA,\bar{d}})^T$ to CMPP. If  a public data  table $T$ containing IA and the same QIA is accessed, CA and IA can be successfully linked and the complete information of a patient $\tilde{x}=(\tilde{x}_{QIA,d'},\tilde{x}_{CA,\bar{d}},\tilde{x}_{IA,\tilde{d}})^T$ is then achieved by attackers, which is referred  
  as the classical attribute attack \citep{sweeney2002k}. Since medical attributes such as blood lipids, blood pressure, and blood sugar can vary across time, conditions, or locations, obtaining completely identical quasi-identifiers (QIAs) is challenging, which necessitates the following $\mu$-attribute attack for CMPP.
\begin{definition}[$\mu$-attribute attack]\label{definition:attribute-attack}
	Let $\mu\geq0$ and an  attacker $\mathcal A$
access  attack samples $T=\{\xi_l\}_{l=1}^L$ with $\xi_l=(\xi_{l,QIA,d'},\xi_{l,IA,\tilde{d}})^T\in\mathbb I^{d'+\tilde{d}}$.  For a patient who submits   $x=({x}_{QIA,d'},{x}_{CA,\bar{d}})^T \in\mathbb I^{d'+\bar{d}}$ to CMPP, define
$ 
    \xi_{x,QIA,d'}:=\xi_{l^*,QIA,d'}
$ 
with $l^*=\arg\min_{l=1,\dots,L} \|x_{QIA,d'}-\xi_{l,QIA,d'}\|_2$, where $\|\cdot\|_2$ denotes the Euclidean norm.
If $  \|x_{QIA,d'}-\xi_{x,QIA,d'}\|_2\leq \mu$, 
then  $x_{QIA,d'}$ is $\mu$-linked to $\xi_{x,QIA,d'}$ and the patient  is $\mu$-attribute attacked in the sense that a complete   $ (\xi_{l^*,IA,\tilde{d}},{x}_{QIA,d'},{x}_{CA,\bar{d}})$ is achieved. 
\end{definition}

To defend against $\mu$-attribute attacks presented in Definition \ref{definition:attribute-attack},
 patients are frequently suggested to submit an anonymized query   ${x}_{QIA,d'}^{\delta}$ satisfying $\|x_{QIA,d'} - {x}_{QIA,d'}^{\delta}\|_2 > 2\mu$ so that  $\|{x}_{QIA,d'}^{\delta} - \xi_{x,QIA,d'}\|_2  >  \mu$, where  $\delta$ is a perturbation parameter. 
The condition $\|x_{QIA,d'} - {x}_{QIA,d'}^{\delta}\|_2 > 2\mu$ therefore indicates immunity to $\mu$-attribute attacks, which follows the following privacy principle. 

\begin{definition}[$2\mu$-Correct Orientation]\label{Def.correct orientation}
	Given a set $\Xi_N:=\{\eta_i\}_{i=1}^N$ with $\eta_i\in\mathbb R^d$, and its perturbed counterpart $\Xi_N^\delta:=\{\eta_i^\delta\}_{i=1}^N$,   $2\mu$-correct orientation (CO) is defined by
	\begin{equation}\label{def.incorrect orientation_p}
	CO(\Xi_N,\Xi_N^{\delta}, \mu) := \frac{\sum_{i=1}^{N}I_{\|\eta_{i}-\eta^{\delta}_{i}\|_2\leq2\mu}}{N} \times 100\%,
	\end{equation}   
	where $I_A$ denotes the indicator on the event $A$.
\end{definition}

\vspace{-0.2cm} 
\begin{figure}[H]
	\centering
	\caption{Illustrative Example of TQMA Perturbation.}
	\vspace{0.1in}
	\includegraphics[scale=0.5]{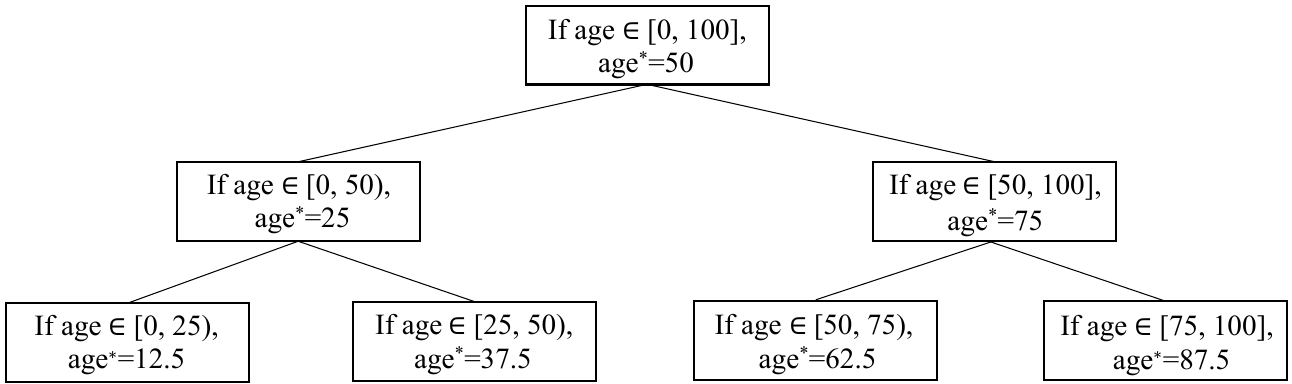}
	\label{Fig:TQMA_tree}  
	\vspace{-0.3cm}
\end{figure}
\vspace{-0.2cm}

According to  Definition \ref{Def.correct orientation}, a smaller CO value on QIA indicates a lower likelihood of $\|x_{QIA,d'} - {x}_{QIA,d'}^{\delta}\|_2 \leq 2\mu$ (or $\|{x}_{QIA,d'}^{\delta} - \xi_{x,QIA,d'}\|_2  <  \mu$), meaning a reduced probability of the patient being vulnerable to $\mu$-attribute attacks.
Consequently, privacy-preserving approaches resulting in smaller CO values are preferable.
To efficiently minimize CO while preserving the original QIA's positional information and providing real-time feedback to patients, we propose a novel method called Tree-based Quasi-Microaggregation (TQMA) that 
utilizes  a pre-constructed binary tree to partition the attribute's value range, all without the need for access to the entire dataset. As shown in  Figure \ref{Fig:TQMA_tree} (or Algorithm 1 in Appendix A), TQMA replaces the original value with the midpoint of  the sub-division interval. The following proposition presents an intuitive  effectiveness verification of TQMA against the $\mu$-attribute attack.

\begin{proposition}\label{prop:TQMA-model}
	Let $v$ be a random variable that follows the uniform distribution on the interval  $[a, b]$ with $a<b$. If TQMA with tree depth $k\in\mathbb N$ is implemented to $v$ to yield a perturbed version $v^{TQMA(k)}$, 
	then for 
	$2\mu\in[0, (b-a)2^{-(k+1)}]$,
	there holds
	\begin{equation}\label{prop: TQMA}
	P(\|v - v^{TQMA(k)}\|_2 \leq 2\mu) \leq \frac{\mu2^{k+2}}{b-a}.
	\end{equation}
\end{proposition}

Proposition \ref{prop:TQMA-model} shows that the tree depth $k$ adjusts the balance between privacy preservation and maintaining position information.
As $k$ increases, the probability that $v^{\text{TQMA}(k)}$ is within $2\mu$ of its original value $v$  increases, while the patient's resistance to $\mu$-attribute attacks  descreases.

\subsection{Privacy Preservation against Model Extraction Attacks}

Given a query point $x$ provided by a patient and a corresponding  response $y=f_v(x)$ made by a victim doctor, 
model extraction attacks happen when  an  attacker gets a model $f_a$ that effectively mimics $f_v$.  As the model of a doctor, even unknown for himself, cannot be actually achieved, we introduce the following $\varepsilon$-model extraction attack \citep{tramer2016stealing} for CMPP. 
 

\begin{definition}[$\varepsilon$-model extraction attack]\label{Def:model-extraction-attack}
	Let $\varepsilon\geq 0$, $\mathbb B$ be a Banach space and $f_v\in\mathbb B$ be the model possessed by a victim $\mathcal V$. If an attacker $\mathcal A$ obtains an approximate model  $f_a\in\mathbb B$ satisfying
	\begin{equation}\label{model-approximation}
	\mbox{dist}_{\mathbb B}(f_a,f_v):=\|f_a-f_v\|_{\mathbb B}\leq \varepsilon,
	\end{equation}
	then the victim is $\varepsilon$-model extraction attacked by $\mathcal A$.
\end{definition}

Assume that the $j$th doctor is attacked and $|D_j|\in \mathbb N$ fake queries  $\Lambda_j^*:=\{x_\ell^{\mbox{fake}}\}_{\ell=1}^{|D_j|}$ are sent to him. 
CMPP consequently collects a set of data $D_j^{\mbox{fake}}:=\{(x_\ell^{\mbox{fake}},f_{D_j, \hat{h}_j}(x_\ell^{\mbox{fake}}))\}_{\ell=1}^{|D_j|}$  over a period of time, where $\hat{h}_j$ is the model parameter of $j$th doctor. CMPP then uses $D_j^{\mbox{fake}}$ to replace the $j$th doctor's local data set and uses the model trained on it to mimic the doctor's decision-making process. It is easy to derive that with $D_j^{\mbox{fake}}$ and the extracted model, CMPP can replace the $j$th doctor without affecting the final synthesized prediction accuracy. This demonstrates that doctors in CMPP are vulnerable to model attraction attacks.

As the execution of model extraction attacks relies heavily on the input–output correspondences.
A preferable strategy to defend against model extraction attacks in CMPP is to perturb the outputs provided by doctors, disrupting the input–output correspondences so that attackers cannot establish a model $f_a$ that achieves $\|f_a-f_v\|_{\mathbb B}\leq \varepsilon$.
Regard $f^{\beta}_{D_j, \hat{h}_j}(x)$  that satisfies  $\|f^{\beta}_{D_j, \hat{h}_j}(x)-f_{D_j, \hat{h}_j}(x)\|_2> \varepsilon$ as the anonymized version of $f_{D_j, \hat{h}_j}(x)$ that cuts off the original input–output correspondence, where $\beta$ denotes a perturbation.
A principle that measures the likelihood of $\|f^{\beta}_{D_j, \hat{h}_j}(x)-f_{D_j, \hat{h}_j}(x)\|_2> \varepsilon$ is therefore needed to assess the $j$th doctor's ability to resist model extraction attacks. 
We then slightly modify the  distance-based record linkage (RL) \citep{Li2006} to measure the quality of privacy preservation in the following definition.

\begin{definition}[distance-based record linkage]\label{def:RL}
Let $\Xi_N:=\{\eta_i\}_{i=1}^N$ and $\Xi_N^\beta:=\{\eta_i^\beta\}_{i=1}^N$ be the sets of original values and their perturbed counterparts, respectively.
For any $\eta^{\beta}_i\in \Xi_N^\beta$,  define $i^*=\arg\min_{1\leq i'\leq N}\|\eta^{\beta}_i-\eta_{i'}\|_2$
and 
$
\mbox{dist}_{2,i}(\eta^{\beta}_i,\Xi_N): 
=\min_{i'\neq i^*}\|\eta^{\beta}_{i}-\eta_{i'}\|_2$.
A record in $\Xi_N^\beta$ is   linked to $\Xi_N^\beta$, if
$$
\|\eta^{\beta}_{i}-\eta_{i}\|_2\leq \mbox{dist}_{2,i}(\eta^{\beta}_i,\Xi_N).
$$
 RL is defined to be   the rate of linked records, 
\begin{equation}\label{Def.RL}
	RL(\Xi_N,\Xi_N^{\beta}):=\frac{\left|\{i:\|\eta^{\beta}_{i}-\eta_{i}\|_2\leq \mbox{dist}_{2,i}(\eta^{\beta}_i,\Xi_N)\}\right|}{N}\times100\%.
\end{equation}   
\end{definition}

According to Definition \ref{def:RL}, a smaller RL value on output indicates a lower likelihood of identifying a doctor's original input–output pairs, thereby reducing the doctor's vulnerability to model extraction attacks. 
To reduce RL and maintain the final synthesized result,
we develop the Bounded Swapping and Threshold Decryption mechanism (BSTD).
BSTD combines bounded swapping \citep{Li2011} and threshold decryption \citep{lindell2005secure}. 
 Bounded swapping is a three-step perturbation approach that disrupts input–output correspondence. Regarding a set of real numbers $\{a_1,\dots,a_m\}$ as the outputs provided by doctors, bounded swapping sets a lower bound $p_\mathrm{lower}$  and an upper bound $p_\mathrm{upper}$ for swapping, ranks the real numbers to obtain a set $\{a_1^*,\dots,a_m^*\}$, where $a_1^*\geq \dots\geq a_m^*$, and randomly selects one from the set $\{a^*_{j- p_\mathrm{upper}},\dots,a^*_{j-p_\mathrm{lower}},a_{j+p_\mathrm{lower}}^*,\dots,a^*_{j+p_\mathrm{upper}}\}\cap\{a_1^*,\dots,a_m^*\}\backslash \{j\}$ to swap with $a_j^*$.  
Threshold decryption such as the $t$-out-of-$\ell$ threshold  scheme sets a restriction on multiparty collaboration; a collaboration  is rejected if fewer than $t$ parties  agree to participate. We set two threshold decryptions with $t=\ell=m$, where $m$ is the number of doctors. 
The first threshold decryption controls entry into BSTD, preventing any $m-1$ doctors from colluding with CMPP. The second manages BSTD's black box permissions, avoiding CMPP snooping on the swapping process. The parameters $p_\mathrm{lower}$  and $p_\mathrm{upper}$ are set considering that doctors with high local estimates prefer exchanges with those whose predictions closely align with theirs. These two parameters address doctors' concerns about fair contribution allocation and keep them informed about the range of swapping, fostering a trustworthy environment. The detailed implementation of BSTD can be found in Algorithm 2 in Appendix A. The following proposition illustrates the effectiveness against the model extraction attacks.

\begin{proposition}\label{prop:BSTD-model}
	If BSTD with $p_\mathrm{lower}, p_\mathrm{upper}\in\mathbb N$ is implemented to the submitted   local outputs $\{f_{D_j}(x^{\mbox{fake}}_\ell)\}_{j=1}^m$ to yield a set
	of swapped outputs $\{f_{D_j}^{p_\mathrm{lower},p_\mathrm{upper}}(x^{\mbox{fake}}_\ell)\}_{j=1}^m$, then for any $j=1,\dots,m$, 
	there holds
	\begin{equation}\label{probability-bound}
	P\left[f_{D_j}^{p_\mathrm{lower},p_\mathrm{upper}}(x_\ell^{\mbox{fake}})=f_{D_j}(x_\ell^{\mbox{fake}})\ \mbox{for all}\ \ell=1,\dots,|D_j|\right]
	\leq
	\frac1{(p_\mathrm{upper}-p_\mathrm{lower}+1)^{|D_j|}}.
	\end{equation}
\end{proposition}

Proposition \ref{prop:BSTD-model} indicates that by increasing $p_\mathrm{upper}$ or decreasing $p_\mathrm{lower}$, BSTD  can reduce the likelihood of $f_{D_j}^{p_\mathrm{lower},p_\mathrm{upper}}(x_\ell^{\mbox{fake}})$ being linked to $f_{D_j}(x_\ell^{\mbox{fake}})$, i.e., reduce the likelihood of the output $f_{D_j}(x^{\mbox{fake}}_\ell)$ being linked to its corresponding input $x^{\mbox{fake}}_\ell$, thereby protecting doctors from input–output correspondence-based model extraction attacks.

\section{TQMA-BSTD-based Distributed Learning Framework for  PPCMPP}\label{Sec:platform_now}
This section focuses on efficiently integrating TQMA, BSTD with a one-shot swapped  distributed learning framework  to guarantee the privacy and accuracy requirements of PPCMPP .  

\subsection{Problem Formulation: From Optimization to Machine Learning}
Assume that the $j$th doctor possesses a data set  
$D_j:=\{(x_{i,j},y_{i,j})\}_{i=1}^{|D_j|}$  with $x_{i,j}\in\mathbb I^d$ being i.i.d. drawn according to an unknown distribution $\rho$ and  $y_i\in\mathcal Y\subset[-M,M]$ for some $M>0$ satisfying 
\begin{equation}\label{data-form}
    y_{i,j} =f^{\diamond}(x_{i,j}) + \epsilon_{i,j},
\end{equation} 
 where $\epsilon_{i,j}$ is independent bounded zero-mean noise, i.e.,  $|\epsilon_i|\leq M$,  and $f^{\diamond}:\mathbb I^d\rightarrow \mathcal Y$ is the ground truth relation between inputs and outputs.    
Given a set of queries $\Xi_N=\{x_i^*\}_{i=1}^N$ for $N\in\mathbb N$, to defend against the attribute attack, the TQMA is implemented to them and  perturbed counterparts $\Xi_N^{TQMA(k)}=\{x_i^{(TQMA(k)}\}_{i=1}^N$ with tree depth $k$  are obtained in CMPP. Fed with a perturbated query, the $j$th doctor submits the response  $f_{D_j}(x_i^{(TQMA(k)})$   to CMPP and the BSTD mechanism is implemented to the response to get a perturbed version $f^{p_\mathrm{lower},p_\mathrm{upper}}_{D_j}(x_i^{(TQMA(k)})$. CMPP then synthesizes the final response 
\begin{equation}\label{target-mechansim}
    \overline{f}_{D}(x_i^*):=\mathcal S(f^{p_\mathrm{lower},p_\mathrm{upper}}_{D_1}(x_i^{(TQMA(k)}),\cdots,f^{p_\mathrm{lower},p_\mathrm{upper}}_{D_m}(x_i^{(TQMA(k)})),
\end{equation}
where $\mathcal S:\mathbb R^m\rightarrow \mathbb R$ is a synthesization mapping.   
Our purpose is then to find an suitable $\mathcal S$ and modification of $f_{D_j}$ provided by $j$th doctor 
to minimize  
$ 
(f^{\diamond}(x)-\overline{f}_{D}(x))^2 
$ 
for any given $x$, CO budget $U>0$ and RL budget $V>0$.

Assume that $\mathcal M$ is a set of functions to encode the a-priori information of $f^\diamond$ and $\Lambda$ is the set of distributions of $\rho$. Given the critical importance of accuracy in medical prediction, we are interested in  the worst-case error, defined by 
{\setlength{\abovedisplayskip}{4pt}
	\setlength{\belowdisplayskip}{3pt}
\begin{align}\label{utility-measure}
\mathcal U_{\mathcal M,\Lambda}(\overline{f}_{D},x):= \sup_{f^{\diamond}\in \mathcal M,\rho\in\Lambda}E[(f^{\diamond}(x)-\overline{f}_{D}(x))^2],\qquad\forall x\in\mathbb I^d.
\end{align}}
The purpose of PPCMPP then boils down to the optimization problem:
{\setlength{\abovedisplayskip}{4pt}
	\setlength{\belowdisplayskip}{3pt}
\begin{equation}\label{P}
\begin{aligned}
& \qquad\inf_{f_D \in \Psi_D} \mathcal{U}_{\mathcal{M}, \Lambda}(f_D, x^{TQMA(k)}), \quad x\in\mathbb I^d\\
& \text{s.t.} \quad CO(\Xi_N, \Xi_N^{TQMA(k)}, \mu) \leq U, \quad i = 1, \dots, N, \\
& \qquad RL\Big(\{f_{D_j}(x_i^{TQMA(k)})\}_{j=1}^m,\ \{f_{D_j}^{p_{\text{lower}}, p_{\text{upper}}}(x_i^{TQMA(k)})\}_{j=1}^m\Big) \leq V, \quad i = 1, \dots, N
\end{aligned}
\end{equation}
where $\Psi_D$ denotes the class of all learning models derived from the dataset $D=\cup_{j=1}^mD_j $.  
 
Since $\Psi_D$ is uncountable and cannot be parameterized, the optimization problem \eqref{P} is unsolvable, implying that it is impossible to obtain a PPCMPP scheme by solving \eqref{P}. We relax the problem \eqref{P} by means  of machine learning, since the infimum problem
$\inf_{f_D\in \Psi_D}\mathcal U_{\mathcal M,\Lambda}(f_D,x_i^*)$ is  theoretically achievable for some one-shot distributed learning equipped with local average regression \citep{chang2017distributed} and kernel methods \citep{lin2017distributed} in the sense of rate optimality. 
To be detailed, though problem \eqref{P} is unsolvable, it is possible  to construct  some  distributed learning schemes framework to obtain $\bar{f_D}$ satisfying  
{\setlength{\abovedisplayskip}{4pt}
	\setlength{\belowdisplayskip}{3pt}
\begin{equation}\label{P-variant}
\begin{aligned}
& \qquad\mathcal{U}_{\mathcal{M},\Lambda}(\bar{f}_D, x^{TQMA(k)} ) \sim \inf_{f_D \in \Psi_D} \mathcal{U}_{\mathcal{M},\Lambda}(f_D, x ),  x\in\mathbb R \\
& \text{s.t.} \quad CO(\Xi_N, \Xi_N^{TQMA(k)}, \mu) \leq U,  \\
& \qquad RL\Big(\{f_j(x_i^{TQMA(k)})\}_{j=1}^m,\ \{f_j^{p_{\text{lower}}, p_{\text{upper}}}(x_i^{TQMA(k)})\}_{j=1}^m\Big) \leq V, \quad i = 1, \dots, N.
\end{aligned}
\end{equation}
We focus on designing one-shot distributed learning framework via appropriate settings of $\mathcal S$ and $f_{D_j}$ so that $\mathcal{U}_{\mathcal{M},\Lambda}(\bar{f}_D, x^{TQMA(k)}) \sim \inf_{f_D \in \Psi_D} \mathcal{U}_{\mathcal{M},\Lambda}(f_D, x)$.

\subsection{One-shot Distributed Learning Framework for PPCMPP}
Presenting a   scheme to determine the synthesization scheme $\mathcal S$ and the local estimator $f_{D_j}$ in \eqref{target-mechansim} to satisfy \eqref{P-variant} is quite difficult since  TQMA and BSTD mechanisms leads to perturbation of both queries and local estimates made by doctors but the prediction accuracy should be still optimal.  In particular, to guarantee that BSTD does not affect the prediction accuracy, $\mathcal S$ should be selected to be symmetric with respect to $f_{D_j}$, making the one-shot non-parametric distributed learning scheme \citep{Zhang2015,lin2017distributed,chang2017distributed} based on divide-and-conquer a preferable approach for this purpose. In addition, to reduce the negative effect of TQMA in prediction, some qualification mechanism \citep{liu2022enabling} should be introduced to measure the quality of local estimates.

Our approach  combines TQMA-BSTD mechanisms with 
a delicate one-shot distributed learning framework and divides into six steps as shown in Figure \ref{platform_pri}.
We start with a TQMA interaction window for patients to select a tree depth $k$ to receive a   perturbed version of query. Then, the CMPP platform evaluates the qualification of the $j$th doctor based on their registration information such as  age, job title, years of work experience, and education to mimic the data size $|D_j|$ the doctor possesses and sends both $|D_j|$ and $|D|=\sum_{j=1}^m|D_j|$ to the $j$th doctor.  In the third step, the $j$th doctor makes the initial response $f_{D_j,h_j}(x^{(TQMA(k)})$ with his own  parameter (or evaluation principle) $h_j\in(0,1)$ and then submits the local estimate  as a scaled version of $f_{D_j,h_j}(x^{(TQMA(k)})$, i.e., $\frac{|D_j|}{|D|}f_{D_j,h_j}(x^{(TQMA(k)})$. This bundled form is hereinafter denoted as $f_{B_j}(x^{(TQMA(k)})$. Different from other one-shot distributed learning systems \citep{Zhang2015,liu2022enabling} that submit the local estimate to the platform directly, our approach needs a BSTD interaction window in the fourth step to swap local estimates $\{f_{B_j}(x^{(TQMA(k)})\}_{j=1}^m$ to $\{f_{B_j}^{p_{lower},p_{upper}}(x^{(TQMA(k)})\}_{j=1}^m$ into,   effectively safeguarding  doctors from model extraction attacks.
In the fifth step, we introduce a final qualification mechanism to determine which doctors are active for presenting non-trivial local estimates to the perturbed queries $x^{TQMA(k)}$. In the final step, a simple addition operator for active  local estimates is suggested as the synthesization method to guarantee the symmetric property of $\mathcal S(\cdot)$. We call the proposed PPCMPP as TQMA-BSTD PPCMPP (TB-PPCMPP) for short, whose detailed implementation is presented in Algorithm 3 in Appendix A.

Besides the  TQMA interaction and BSTD interaction windows  that make  the TQMA–BSTD mechanism   transparency and enable  patients and doctors to clearly understand the relationship between their preservation levels and their chosen privacy parameters,  there are mainly two novelties in the proposed TB-PPCMPP in local estimates construction and active doctors qualification. Due to the different diagnosis strategies of different doctors, our distributed learning scheme adapts to heterogeneous local estimates,  i.e., $f_{D_j,h_j}, j=1,\dots,m$ can be derived by different algorithms with adaptively selected parameters. In TB-PPCMPP, we require the doctor to present more conservative prediction in terms of using smaller $h_j$ than that in their sole prediction. In particular, we borrow the idea as logarithmic mechanism $\hat{h}_j=h_j^{\log_{|D_j|}{|D|}}$ from \citep{liu2022enabling} for  conservative prediction. 
We also establish an active rule, denoted as $|f_{B_j}^{p_{lower},p_{upper}}(x)|  \geq \frac{|D_j|}{|D|^2}$, as the qualification mechanism to exclude exceedingly small local predictions caused by the TQMA perturbation,
ensuring doctors with negligibly contributions do not have equal influence in the synthesis. The threshold $\frac{|D_j|}{|D|^2}$ in qualification is primarily for the purpose of theoretical analysis. 
Denote by  $D_j^*$   the dataset concerning the  $j$th active doctor, $D^*=\bigcup_{j=1}^{m^*}D^*_j$, and $m^*$ is the number of all active doctors.
The  final prediction made by TB-PPCMPP is 
\begin{equation}\label{synthesis_form_active}
\tilde{f}_{D}(x)= \sum_{j=1}^{m^*}\frac{|D|f_{B_j}^{p_{lower},p_{upper}}(x^{TQMA(k)})}{|D^*|}.
\end{equation}

 \vspace{-0.2cm} 
\begin{figure}[H]
	\centering
	\caption{Privacy-Preserving Collaborative Medical Prediction Platform.  }
	\vspace{0.1in}
	\includegraphics[scale=0.45]{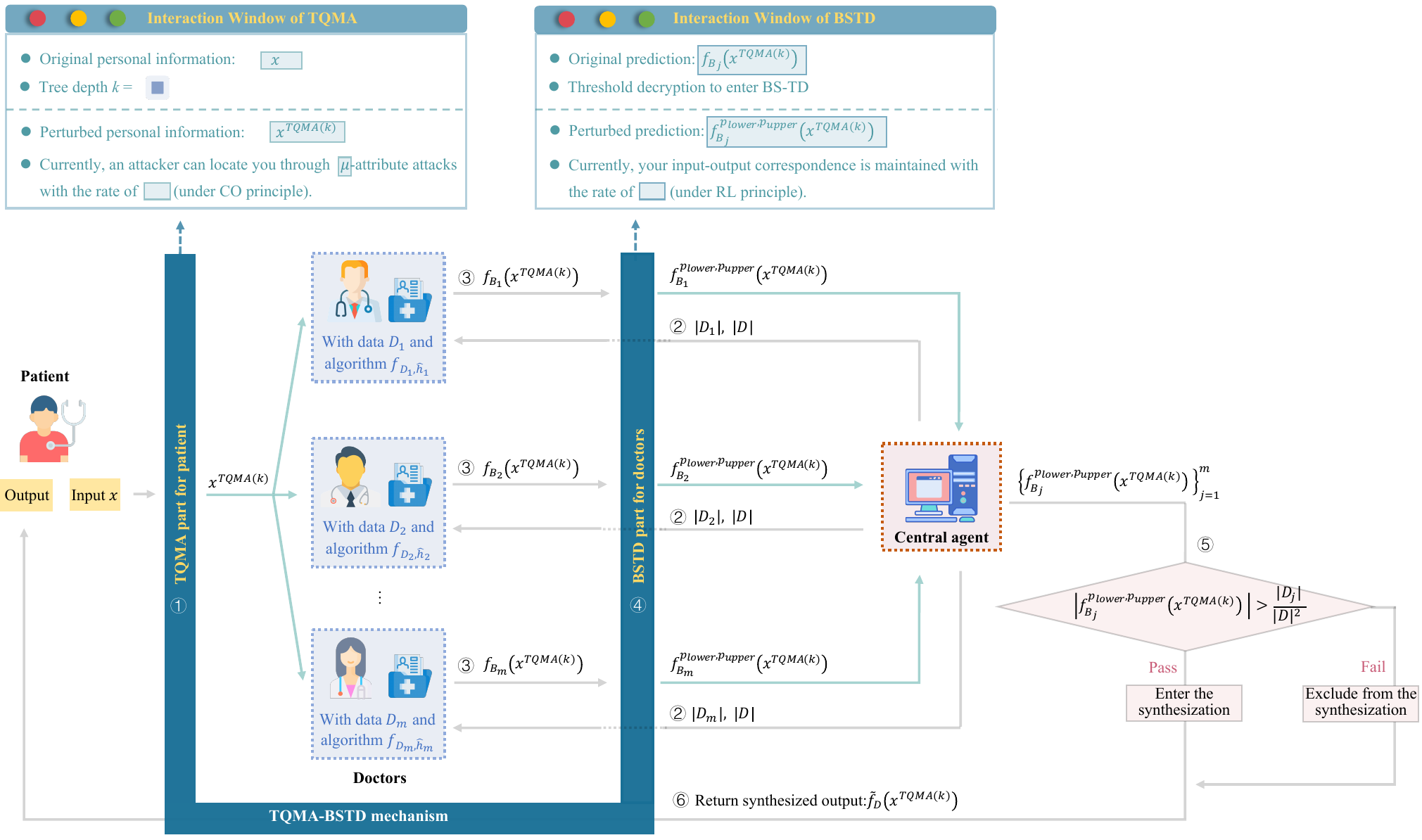} 
	\label{platform_pri}  
	\vspace{-0.1in}
\end{figure}
\vspace{-0.4cm}

\section{Theoretical Verifications}\label{Sec.theory}

This section provides theoretical verifications of the  estimate \eqref{synthesis_form_active} derived from the proposed TB-PPCMPP satisfies \eqref{P-variant}. 
Given the focus on medical prediction, we favor a learning paradigm that mirrors doctors' decision-making, where decisions for new patients are based on similar past cases—known as ``patient similarity-based modeling'' \citep{ng2015personalized, chawla2013bringing}. 
This process can be effectively simulated by local average regression (LAR) \citep{Gyorfi2002}, which selects neighboring data points and calculates weighted averages of their outputs to produce a response. To be specific,   the prediction of the $j$th doctor can be mimicked by:  
\begin{equation}\label{local-estimation}
f_{D_j,h_j}(x)=\sum_{i=1}^{|D_j|}W_{j,h_j,x_{i,j}}(x)y_{i,j},
\end{equation}
where $W_{j,h_j,x_{i,j}}(x)$ is a nonnegative weight that decreases as $x_{i,j}$ moves away from $x$, with $h_j$ measuring similarity between them. Table \ref{Tab:application1} lists common weights, their algorithms, and applications in medical prediction. This follows the following assumption.

\vspace{-0.1cm}
\begin{table}[H]
    \caption{Local Average Regression Algorithms \citep{liu2022enabling}}\label{Tab:application1}
    \vspace{0.1cm}
    \renewcommand\arraystretch{0.75}
    \small
    \begin{center}
        \scalebox{0.88}{ 
        \begin{tabular}{lcl}
            \toprule
            \textbf{Approach} & $W_{j,h_j,x_{i,j}}(x)$ & \textbf{Applications} \\
            \midrule
            NWK (Gaussian) & $\frac{
                \exp\{-\|x-x_{i,j}\|^2/{h_j^2}\}}{\sum_{i=1}^{|D_j|}\exp\{-\|x-x_{i,j}\|^2/{h_j^2}\}}$ & Coronary lumen segmentation \citep{kuncheva2001decision} \\
            \midrule
            NWK (Laplace) & $\frac{
                \exp\{-\|x-x_{i,j}\|/{h_j}\}}{\sum_{i=1}^{|D_j|}\exp\{-\|x-x_{i,j}\|/{h_j}\}}$ & Medical image denoising \citep{mingliang2016medical} \\
            \midrule
            NWK (Epanechnikov) & $\frac{
                (1-\|x-x_{i,j}\|^2/h_j^2)_+}{\sum_{i=1}^{|D_j|}( 1-\|x-x_{i,j}\|^2 /h_j^2)}$ & Death hazard rate estimation \citep{soltanian2012non} \\
            \midrule
            PE & $\frac{
                I_{x_{i,j}\in A_{h_j}(x)}}{\sum_{i=1}^{|D_j|}I_{x_{i,j}\in A_{h_j}(x)}}$ & Medical cost prediction \citep{bang2000estimating} \\
            \midrule
            KNN & $\frac1{k_j}I_{x_{i,j}\in \{x_{(1)},\dots,x_{(k_j)}\}}$ & Kidney discard prediction \citep{barah2021predicting} \\
            \bottomrule
        \end{tabular}}
    \end{center}
    \vspace{0.2cm}
\end{table}
\vspace{-1.0cm}

\begin{assumption}\label{Assumption:local-average}
Assume that each doctor in TB-PPCMPP  produces the prediction   as \eqref{local-estimation} with weights selected from Table \ref{Tab:application1}.
\end{assumption}

Based on Assumption \ref{Assumption:local-average}, a smoothness assumption  on the ground-truth $f^\diamond$ to show that $x\approx x'$ implies $f^\diamond(x)\approx f^\diamond(x')$ and a boundedness assumption of the distribution $\rho$
are necessary, requiring the following assumption that has been widely adopted  in the literature \citep{Gyorfi2002,belkin2019,liu2022enabling}.

\begin{assumption}\label{Assumption:distribution}
	For $0<r\leq 1$, $c_0>0$, $p_{\min},p_{\max}>0$, assume that   $f^\diamond$ satisfies
 \begin{equation}\label{lip}
|f^\diamond(x)-f^\diamond(x')|\leq c_0\|x-x'\|^r,
\end{equation}
for $c_0,r>0$   
    and
	$p_{\min}\leq \rho(x)\leq p_{\max}$ for all $x$ on its support.
\end{assumption}


Denote by $\mathcal M^{r,c_0}$ and $\Lambda_{p_{\min},p_{\max}}$ the set of all $f^\diamond$ and $\rho$ in Assumption \ref{Assumption:distribution}, respectively. 
 It can be found in   \citep{Gyorfi2002,liu2022enabling} that for any $x\in\mathbb I^d$, there holds
\begin{equation}\label{best}
 \inf_{f_D \in \Psi_D} \mathcal{U}_{\mathcal{M}^{r,c_0},\Lambda_{p_{\min},p_{\max}}}(f_D, x ) \sim |D|^{-2r/(2r+d)}.
\end{equation}
 We rigorously prove in the following theorem that the derived estimate in \eqref{synthesis_form_active} satisfies \eqref{P-variant}.

\begin{theorem}\label{Theorem:Optimal}
	Let $\{x_i^*\}_{i=1}^N$ be the set of queries,  $\hat{h}_j=h_j^{\log_{|D_j|}|D|}$, $x^{TQMA(k)}$ be a perturbed version of $x\in\mathbb I^d$ via TQMA  with tree depth $k\in\mathbb N$, and $\tilde{f}_{D}(x^{TQMA(k)})$ be defined by \eqref{synthesis_form_active}   with $p_\mathrm{lower} \geq 2$ and $W_{j,h_j,x_{i,j}}(x)$ being given in Table \ref{Tab:application1}.  If Assumption \ref{Assumption:local-average} and Assumption \ref{Assumption:distribution} hold, $|D_1|\sim\cdots\sim|D_m|$, $h_j\sim |D_j|^{-1/(2r+d)}$ and there exists at least one $j$ satisfying $|f_{D_j,\hat{h}_j}(x^{TQMA(k)})|\geq|D|^{-1}$ with some $k\geq \frac{\log_2|D|}{4r+2d}-1$,then  
\begin{equation}\label{CO_RL:bound}
\begin{aligned}
&  
 C_1 |D|^{-\frac{2r}{2r+d}}
 \leq \inf_{f_D \in \Psi_D} \mathcal{U}_{\mathcal{M},\Lambda}(f_D, x )\leq \mathcal U_{\mathcal M^{r,c_0},\Lambda_{p_{\min},p_{\max}}}
(\tilde{f}_D, x^{TQMA(k)}) \leq  C_2|D|^{-\frac{2r}{2r+d}} \log^{2r}\!\!\!|D|, \\
& \text{s.t.} \quad CO(\Xi_N, \Xi_N^{TQMA(k)}, \mu) \leq \frac{  2^{k+2}\mu p_{\max}}{b-a}\%, \quad i = 1, \dots, N\\
& \qquad RL\Big(\{f_{B_j}(x_i^{TQMA(k)})\}_{j=1}^m,\ \{f_{B_j}^{p_{\text{lower}}, p_{\text{upper}}}(x_i^{TQMA(k)})\}_{j=1}^m\Big) \leq \frac{100(p_\mathrm{lower}-1)}{m}\%,  
\end{aligned}
\end{equation} 
where  $\mu\!\in\![0, \!(b\!-\!a)2^{-(k+2)}\!]$, $C_1,C_2$ are   constants independent of $|D_j|$. 	
\end{theorem}


It should be highlighted (in Appendix D) that the logarithmic term in \eqref{CO_RL:bound} is removable if NWK (Gaussian) and NWK (Laplace) from Table~\ref{Tab:application1} are excluded. As a consequence,
Theorem \ref{Theorem:Optimal}   confirms that  the  machine learning problem~\eqref{P-variant} can be  solved when $U \geq \frac{  2^{k+2}\mu p_{\max}}{b-a}\%$ and $V \geq \frac{100(p_\mathrm{lower}-1)}{m}\%$. Based on  Theorem \ref{Theorem:Optimal}, under TB-PPCMPP, patients can choose the smallest $k$ satisfying $k\geq \frac{\log_2|D|}{4r+2d}-1$, and doctors can set the minimum $p_\mathrm{lower}$ value to achieve the highest level of privacy preservation without losing prediction accuracy.

In contrast to previous theoretical studies \citep{Li2006,Li2011,dwork2008differential, chaudhuri2011differentially}   that preserving privacy often had a pronounced negative impact on prediction accuracy, our study represents a pioneering effort, to the best of our knowledge, to  develop  a practical preservation mechanism to guarantee high level of preservation without compromising accuracy.

\section{Numerical Verifications}\label{experiment}

We conduct toy simulations and a real-world data experiment to demonstrate the effectiveness of TB-PPCMPP in preserving the privacy of both patients and doctors without compromising prediction performance.  Experimental settings are provided in Appendix A.
Table~\ref{tab:metrics} summarizes the symbols used in the experiments and their meanings.

\vspace{-0.5cm}
\begin{table}[H]
\centering
\caption{Summary of symbols and their meanings used in experiments}
\label{tab:metrics}
\small
\renewcommand\arraystretch{1.3}
\vspace{0.2cm}
\scalebox{0.83}{ 
\begin{tabular}{ll|ll}
  \toprule
\textbf{Symbol} & \textbf{Meaning} & \textbf{Symbol} & \textbf{Meaning} \\
\hline
AE & Avg. error of CMPP/PPCMPP &  $CO_\mathrm{ori}$& 
CO of patients' original inputs  \\
$AE_\mathrm{ori}$ &AE of CMPP without preservation  & $CO_\mathrm{Tk}$ & 
{\makecell[l]{CO under TQMA ($k$) }}   \\
$AE_\mathrm{Tk}$ & AE of PPCMPP equipped with TQMA ($k$)
&  $RL_\mathrm{ori}$ & RL of doctors' original outputs \\
$AE_\mathrm{TkBp_{\mathrm{lower}} p_{\mathrm{upper}}}$ & AE under TQMA ($k$) and BSTD ($p_{\mathrm{lower}}$, $p_{\mathrm{upper}}$)  & $RL_\mathrm{Bp_{\mathrm{lower}} p_{\mathrm{upper}}}$ &  {\makecell[l]{RL under BSTD ($p_{\mathrm{lower}}$, $p_{\mathrm{upper}}$) }}    \\
 $AE_\mathrm{Np_\mathrm{noise}Bp_{\mathrm{lower}} p_{\mathrm{upper}}}$ & AE under Noise ($p_\mathrm{noise}$)  and BSTD ($p_{\mathrm{lower}}$, $p_{\mathrm{upper}}$) & $RL_\mathrm{TkBp_{\mathrm{lower}} p_{\mathrm{upper}}}$ & RL under TQMA ($k$) and BSTD ($p_{\mathrm{lower}}$, $p_{\mathrm{upper}}$)  \\
  $AE_\mathrm{TkNp_\mathrm{noise}}$ &  AE under TQMA ($k$) and Noise ($p_\mathrm{noise}$) 
  & $\bullet\circ$-PPCMPP & PPCMPP with preservation methods $\bullet$ and $\circ$   \\ 
    \bottomrule
\end{tabular}}
\vspace{3.5pt} 
        \footnotesize
        \begin{tabular}{@{}l@{}}
            \multicolumn{1}{@{}p{0.99\textwidth}@{}}{
                \scriptsize
                \textbf{Note}: TQMA ($k$) refers to TQMA with tree depth $k$, BSTD ($p_{\mathrm{lower}}$, $p_{\mathrm{upper}}$) refers to BSTD with parameters $p_{\mathrm{lower}}$ and $p_{\mathrm{upper}}$, and Noise ($p_\mathrm{noise}$) means noising method with privacy parameter $p_\mathrm{noise}$.
            }
        \end{tabular}
\end{table}

\subsection{Toy Simulations}\label{simulation}

We design three toy simulations. 
In the first simulation, we evaluate the performance of TQMA in defending against attribute attacks and highlight its advantages by comparing it with other widely used privacy-preserving approaches. In the second simulation, we assess the effectiveness of BSTD in defending against model extraction attacks and highlight its advantage in balancing the privacy and prediction compared to the noising method. In the third simulation, we evaluate TB-PPCMPP's effectiveness in preserving the privacy for both patients and doctors, and then demonstrate its advantages in leveraging TQMA-BSTD as the privacy-preserving mechanism. Specifically, we combine TQMA, BSTD, and noising methods to create four variants: TB-PPCMPP, TN-PPCMPP, NB-PPCMPP, and NN-PPCMPP, where “N” refers to a noising method. 

The toy simulation employs two noising methods: (1) \textit{Mul}$_{p_{\text{noise}}}$-Noise: multiplicative noise \citep{adam1989security}, defined as $x^* = x \times e$, where $e$ has a mean of zero and a variance of $p_{\text{noise}} \sigma_x^2$. $\sigma_x^2$ denotes the variance of the original data $x$, and $p_{\text{noise}}$ controls the privacy level; and (2) \textit{DP}$_{\epsilon}$-Noise: Laplace noise from $\epsilon$-differential privacy \citep{dwork2008differential}, where the sensitivity $s$ is computed as $s = \max_{j} |f_{B_j}(x_i)|$ on the doctor side and $s = \max_i|(x_i)|$  on the patient side.

We generate training samples $\{(x_{i,j}, y_{i,j})\}_{i=1}^{|D_j|}$ as the data held by the $j$th doctors, with $j=1,\dots, m$ and testing samples $\{(x_i^*, y_i^*\}_{i=1}^{N'}$. 
$x_{i,j}$ and $x_{i}^{*}$ are drawn i.i.d. from the (hyper)cube $[0,1]^{5}$ according to the uniform distribution. 
 $y_{i,j} = f^{\diamond}(x_{i,j}) +\epsilon_{i,j}$ and $y_{i}^{*}=f^{\diamond}(x_i^*)$, where $\epsilon_{i, j}$ is the Gaussian noise $\mathcal{N}(0,0.1)$ and
$$f^{\diamond}(x)=\left\{\begin{array}{cc}(1-\|x\|)_{+}^{5}(1+5\|x\|)+\frac{1}{5}\|x\|^{2}, & 0<\|x\| \leq 1, x \in \mathbb{R}^{5}, \\\frac{1}{5}\|x\|^{2},& \text{otherwise}.\end{array}\right.$$
Let $D=\bigcup_{j=1}^{m}D_j$ with $D_j \cap D_{j'} = \varnothing$ for $j\neq j'$, and set $|D|=10,000$, $N'=1,000$, $m=20$, and $\mu=10^{-3}$ in $\mu$-attribute attacks.

\subsubsection{Effectiveness of TQMA}

 This simulation demonstrates the performance of TQMA in terms of both privacy preservation and prediction performance. As shown in 
Figure \ref{fig:TQMA_effectiveness}, $AE_\mathrm{T4}$ is extremely close to $AE_\mathrm{ori}$, with only a 1.92\% change, while CO drops significantly from 100.00\% to 6.37\%, meaning that up to 936 out of 1,000 patients are immune to the current $\mu$-attribute attacks. This shows the effectiveness of TQMA with a suitable tree depth in resisting $\mu$-attribute attacks without sacrificing prediction performance, thereby justifying Theorem \ref{Theorem:Optimal}.

\vspace{-0.3cm} 
\begin{figure}[H]
	\centering
	\caption{Effectiveness of TQMA.}
	\subfigure{\includegraphics[scale=0.30]{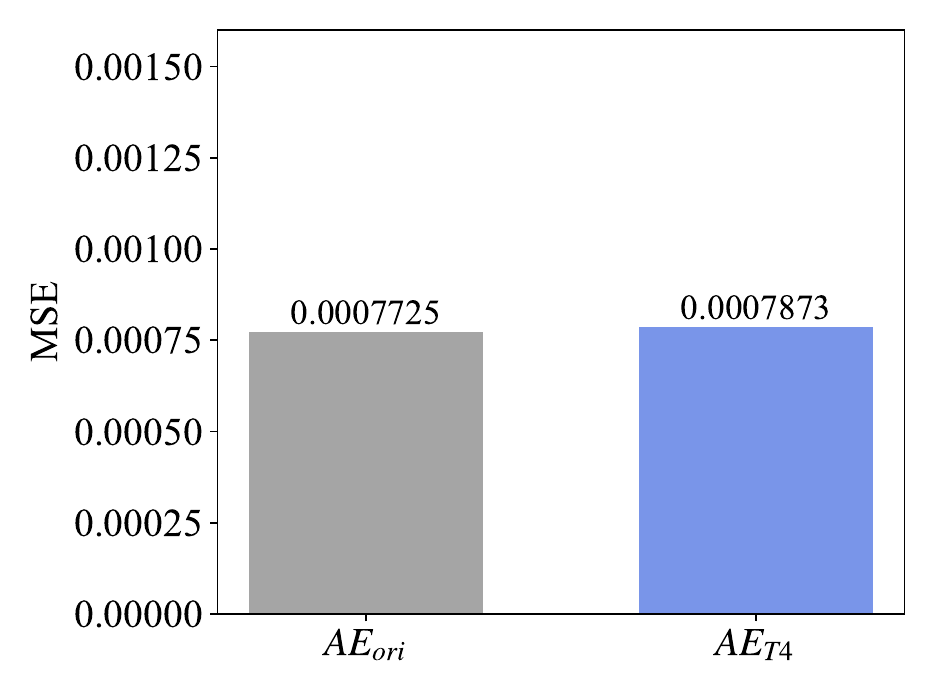}
		\label{subfig: TQMA_MSE}}
	\subfigure{\includegraphics[scale=0.30]{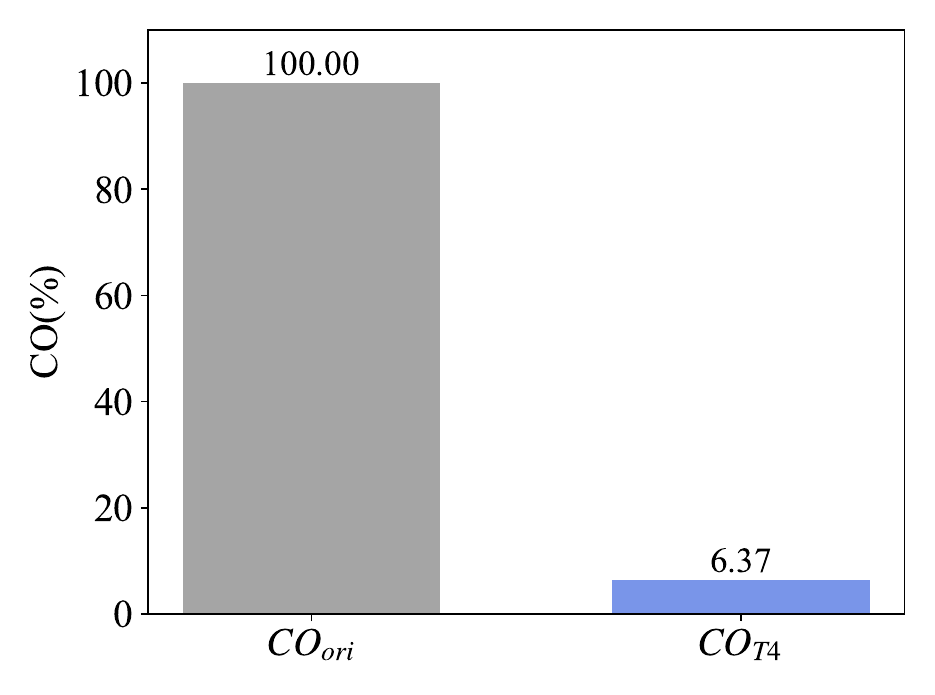}
		\label{subfig: TQMA_CO}}
			\label{fig:TQMA_effectiveness}
\end{figure}
\vspace{-0.8cm} 

We then compare TQMA with other privacy-preserving approaches to demonstrate its advantages. CO is used as the control variable to make the comparison feasible. Specifically, the privacy parameters, namely, the tree depth $k$ of TQMA, the number of groups of microaggregation \citep{domingo2002practical}, the pre-specified leaf size of kd-tree perturbation \citep{Li2006}, the $p_{\text{noise}}$ in \textit{Mul}$_{p_{\text{noise}}}$-Noise, and the $\epsilon$ in \textit{DP}$_{\epsilon}$-Noise, are adjusted so that TQMA offers the strongest defense against $\mu$-attribute attacks.
As shown in Table \ref{Tab:advantage}, TQMA produces the smallest CO and the comparable RMSE to others,
justifying its ability in guaranteeing both privacy and  accuracy. 
Table~\ref{Tab:advantage} also highlights that TQMA offers real-time feedback: unlike other approaches that rely on group-level information and require patients to wait until a total of $N'$ patients are available, TQMA enables patients to receive immediate responses without delay.

\vspace{-0.2cm}
\begin{table}[H]
    \caption{Advantages of TQMA Perturbation}\label{Tab:advantage}
    \vspace{0.1cm}
    \renewcommand\arraystretch{1}
    \footnotesize
    \begin{center}
        \begin{tabular*}{0.85\linewidth}{@{\extracolsep{\fill}}lccc@{}}
            \toprule
            \textbf{Perturbation Approach} & \textbf{CO (\%)} & \textbf{RMSE (original: 0.02779)} & \textbf{Max waiting time per patient} \\
            \midrule
            \textbf{TQMA} & \textbf{6.37} & \textbf{0.02806} & \textit{\textbf{0t}} \\
            UMA & 6.55 & 0.02807 & $(N'-1)t$ \\
            kd-tree Perturbation & 6.84 & 0.03534 & $(N'-1)t$ \\
            \textit{Mul}$_{p_{\text{noise}}}$-Noise & 6.51 & 0.03202 & $(N'-1)t$ \\
            \textit{DP}$_{\epsilon}$-Noise & 6.40  & 0.02876 & $(N'-1)t$ \\            
            \bottomrule
        \end{tabular*}
        \vspace{3.5pt} 
        \footnotesize
        \begin{tabular}{@{}l@{}}
            \multicolumn{1}{@{}p{0.85\textwidth}@{}}{
                \scriptsize
                \textbf{Note}:  Assuming patients arrive at fixed time intervals of $t$, the maximum waiting time for a patient is calculated as follows: For TQMA, the waiting time is $0t$, as it performs privacy operations immediately for each patient without relying on group data. For other methods, the waiting time is $(N'-1)t$, as they require a dataset of $N'=1000$ patients to initiate the operation.
            }
        \end{tabular}
    \end{center}
\end{table}
\vspace{-0.9cm}

\subsubsection{Effectiveness of BSTD}

In this simulation, we compare the performance of three strategies — no privacy operation, applying BSTD, and adding noise to the doctor side in CMPP — under both non-attack and model extraction attack scenarios. The patient data available to the doctor is considered in two forms: original data and TQMA-perturbed data. 
To ensure a meaningful comparison, we set the privacy parameters such that BSTD consistently provides a higher level of privacy preservation (i.e., a lower RL value) compared to the noising methods. The results are presented in Table~\ref{tab:attack_comparison}.

 Table~\ref{tab:attack_comparison} yields three key observations:
(1) Regardless of whether TQMA was applied to the patient side, the BSTD achieved prediction performance almost identical to that of CMPP without doctor-side privacy operation (i.e., Original CMPP and CMPP with TQMA)), as highlighted in bold black font. This demonstrates the effectiveness of BSTD in maintaining prediction performance.
(2) Under model extraction attacks, applying BSTD significantly degraded the CMPP's prediction performance, as highlighted in bold
blue font, indicating its effectiveness in defending against such attacks.
(3) For the noising methods, although they exhibited resistance to model extraction attacks, their prediction performance remained relatively poor even without attacks. Moreover, when their privacy parameters were tuned to achieve a higher level of privacy, it came at a substantial cost to prediction performance, as indicated by the bold red font. These results highlight the advantage of the BSTD approach over the noising methods, as it achieves both high-level privacy preservation and high prediction accuracy.

\vspace{-0.4cm} 
\begin{table}[H]
\centering
\caption{Comparison of the prediction performance of privacy-preserving approaches}
\label{tab:attack_comparison}
\renewcommand\arraystretch{0.87}
\scalebox{0.65}{
\begin{tabular}{llcc|llcc}
\toprule
Method Type & Method & \multicolumn{2}{c|}{Original Patient Data} & & \multicolumn{2}{c}{TQMA-perturbed Patient Data} \\
 &  & \textit{No attack} & \textit{Attack} & & \textit{No attack} & \textit{Attack}  \\
\midrule
& {\makecell[c]{Original CMPP}} & \textbf{0.0007725} & 0.0008276 &{\makecell[c]{CMPP with TQMA}}& \textbf{0.0007725} & 0.0007873 \\
\midrule
\multirow{3}{*}{\textit{BSTD}}
& {\makecell[c]{BSTD$_{(2,10)}$ (RL=1.12\%)}} & \textbf{0.0007716} & \textcolor{blue}{\textbf{0.0020623}} &{\makecell[c]{BSTD$_{(2,10)}$ (RL= 1.10\%)}}& \textbf{0.0007868} & \textcolor{blue}{\textbf{0.0020792}} \\
& {\makecell[c]{BSTD$_{(3,8)}$ (RL=1.67\%)}} & \textbf{0.0007722} & \textcolor{blue}{\textbf{0.0095931}} &{\makecell[c]{BSTD$_{(3, 8)}$ (RL=1.68\%)}}& \textbf{0.0007869} & \textcolor{blue}{\textbf{0.0023629}} \\
& {\makecell[c]{BSTD$_{(4,10)}$ (RL=1.84\%)}} & \textbf{0.0007869} & \textcolor{blue}{\textbf{0.0021101}} &{\makecell[c]{BSTD$_{(4,10)}$ (RL= 1.81\%)}}& \textbf{0.0007869} & \textcolor{blue}{\textbf{0.0021272}} \\
\midrule
\multirow{4}{*}{\textit{Noising}}
& {\makecell[c]{\textit{Mul$_{36}$}-Noise (RL=10.34\%)}} & 0.0184534 & 0.0219932 &{\makecell[c]{\textit{Mul$_{36}$}-Noise (RL=10.36\%)}} & 0.0184244 & 0.0216806 \\
& {\makecell[c]{\textit{Mul$_{126}$}-Noise (RL=9.20\%)}} & \textcolor{red}{\textbf{0.0460226}} & 0.0546607 &{\makecell[c]{\textit{Mul$_{126}$}-Noise (RL=9.20\%)}} & \textcolor{red}{\textbf{0.0459489}} & 0.0538498 \\
& {\makecell[c]{\textit{DP$_{1.0}$}-Noise (RL=12.39\%)}} & 0.0871015$^\star$ & 0.0187932 &{\makecell[c]{\textit{DP$_{1.0}$}-Noise (RL=12.49\%)}}& 0.0869442$^\star$ & 0.0186720 \\
& {\makecell[c]{\textit{DP$_{0.4}$}-Noise (RL=10.95\%)}} & \textcolor{red}{\textbf{0.5390197}$^\star$} & 0.1114563 &{\makecell[c]{\textit{DP$_{0.4}$}-Noise (RL=10.95\%)}}& \textcolor{red}{\textbf{0.5385243}$^\star$} & 0.1111051 \\
\bottomrule
\end{tabular}}
\vspace{3.5pt} 
\footnotesize
\begin{tabular}{@{}l@{}}
    \multicolumn{1}{@{}p{0.87\textwidth}@{}}{
        \scriptsize
        \textbf{Note}: The subscript ${(2,10)}$ in BSTD$_{(2,10)}$ indicates the BSTD parameters $p_\mathrm{lower} = 2$ and $p_\mathrm{upper} = 10$.
The subscript in \textit{DP}$_{1.0}$-Noise represents the privacy parameter $\epsilon$ in $\epsilon$-differential privacy.
The subscript in \textit{Mul}$_{36}$-Noise denotes the  parameter $p_{\text{noise}}$ in multiplicative noise. We provide an explanation in the Appendix C for the phenomenon that after applying \textit{DP}$_{\epsilon}$-Noise, the prediction performance under model extraction attacks was unexpectedly better than without the attack (see the $\star$-marked results).
    }
\end{tabular}
\end{table}
\vspace{-0.8cm}


\subsubsection{Effectiveness of TB-PPCMPP}
This simulation demonstrates the performance of TB-PPCMPP. As shown in Figure~\ref{fig:effective_CMPT}, $AE_\mathrm{T4B38}$ is nearly identical to $AE_\mathrm{ori}$, with only a negligible 1.86\% change. The CO drops to 6.37\% and RL to 1.68\%, indicating that patients face only a 6.37\% risk of $\mu$-attribute attacks and no doctor (20 doctors with $20 \times 1.68\% < 1$) is vulnerable to model extraction attacks. These results demonstrate the effectiveness of TB-PPCMPP in preserving privacy for both patients and doctors without compromising prediction accuracy, thereby supporting Theorem~\ref{Theorem:Optimal}.

We also conduct two experiments comparing various privacy-preserving operations to demonstrate the advantages of TB-PPCMPP — specifically, its sensitivity to privacy parameters and its superiority in balancing privacy and prediction.

In the first experiment, we compare TN-PPCMPP, NB-PPCMPP, and TB-PPCMPP, where “N” denotes multiplicative noise. As shown in Figures~\ref{fig:power_of_PPCMP}(a), \ref{fig:power_of_PPCMP}(b), and \ref{fig:power_of_PPCMP}(c), we observe that the prediction performance of TB-PPCMPP remains stable across its privacy parameters $k$ and $p_\mathrm{lower}$, while the performance of the other two consistently deteriorates as the noise level increases. This highlights the stability of TB-PPCMPP with respect to privacy parameter settings.

\vspace{-0.3cm} 
\begin{figure}[H]
	\centering
	\caption{Effectiveness of TB-PPCMPP.} 	\label{fig:effective_CMPT}
	\setlength{\subfigcapskip}{-1.2em}
	\subfigure{\includegraphics[scale=0.31]{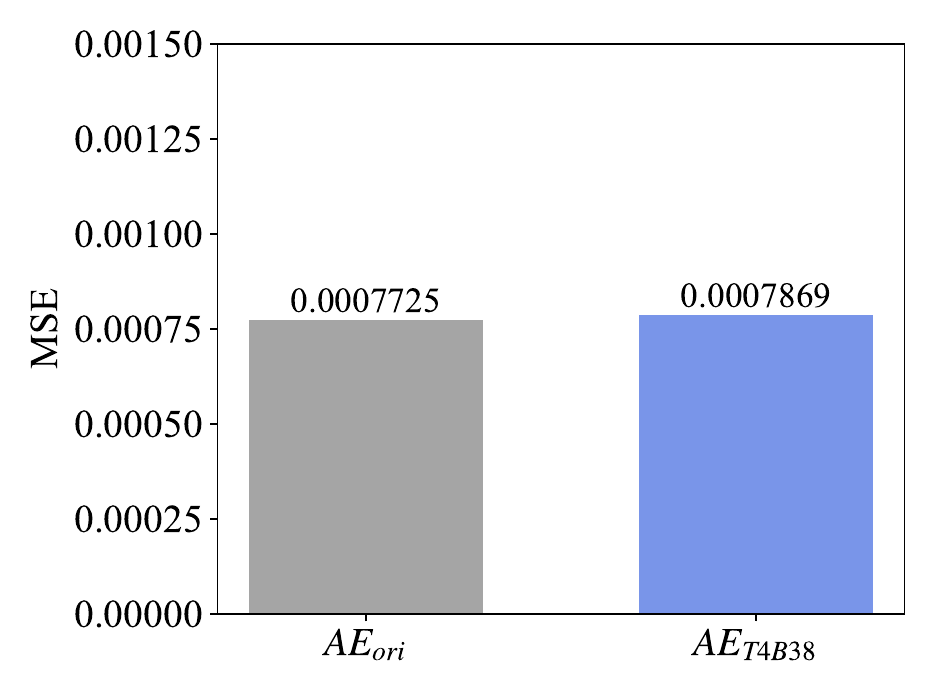}
		\label{subfig: CMPT_MSE}} 
	\setlength{\subfigcapskip}{-1.2em}
	 \subfigure{\includegraphics[scale=0.31]{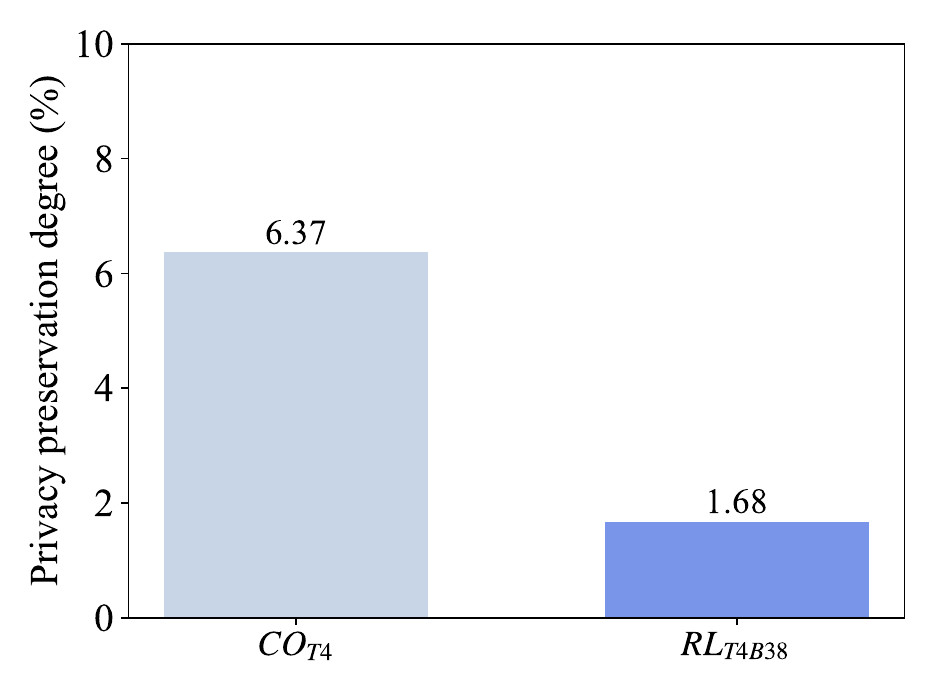}
		\label{subfig: CMPT_CO_RL}} 
\end{figure}
\vspace{-0.9cm}

\begin{figure}[H]
	\centering
	\caption{Advantages of TB-PPCMPP.}
    	\label{fig:power_of_PPCMP}
\setlength{\subfigcapskip}{-1.2em}  
\subfigure[]{\includegraphics[scale=0.32]{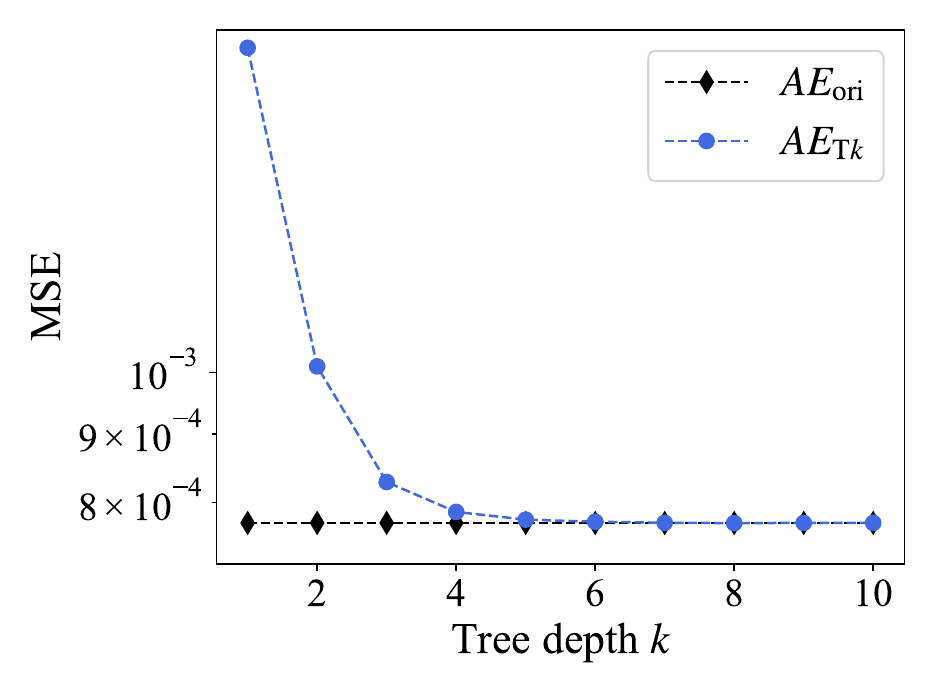}
\label{subfig: TQMA_k_1}}     
\setlength{\subfigcapskip}{-1.2em}
\subfigure[]{\includegraphics[scale=0.32]{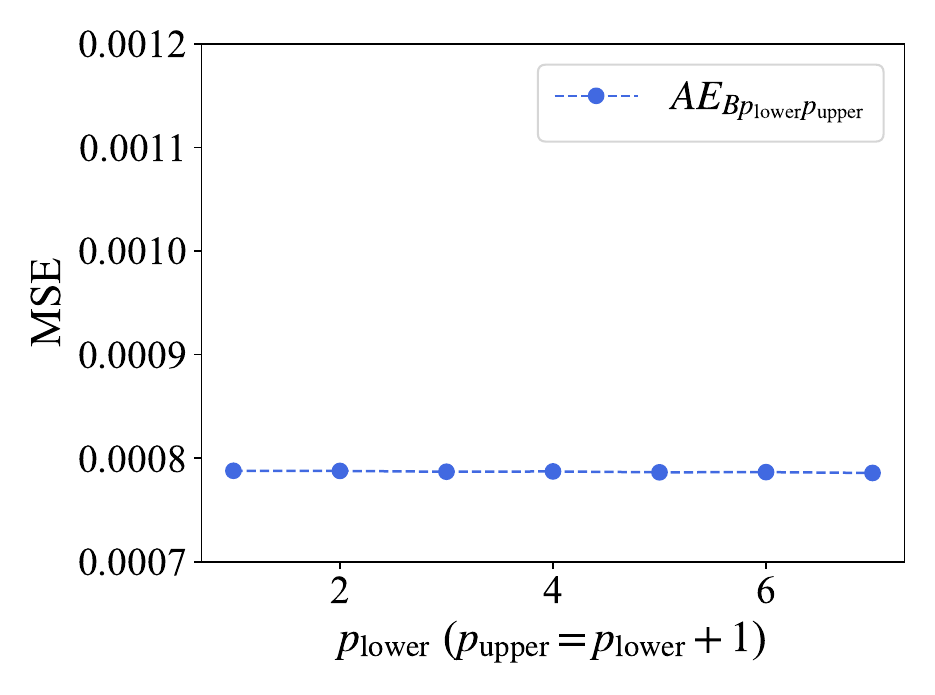}\label{subfig: RL_dis1_AE}} 
\setlength{\subfigcapskip}{-1.2em}
\subfigure[]{\includegraphics[scale=0.32]{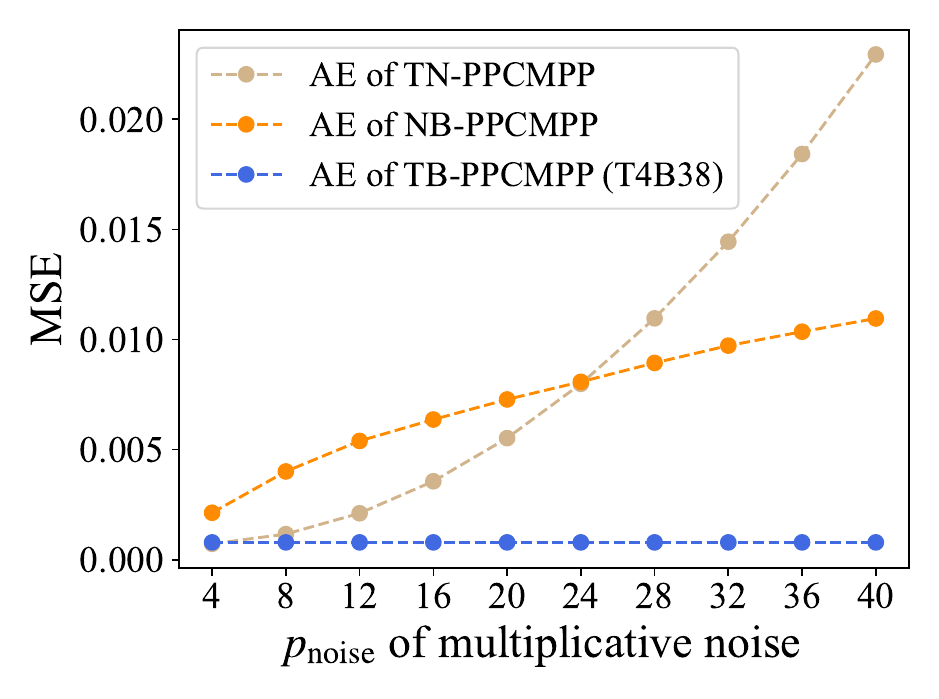}
\label{subfig1}} \\[-0.8em]
\setlength{\subfigcapskip}{-1.2em}
\subfigure[]{\hspace*{0.35em}\includegraphics[scale=0.35]{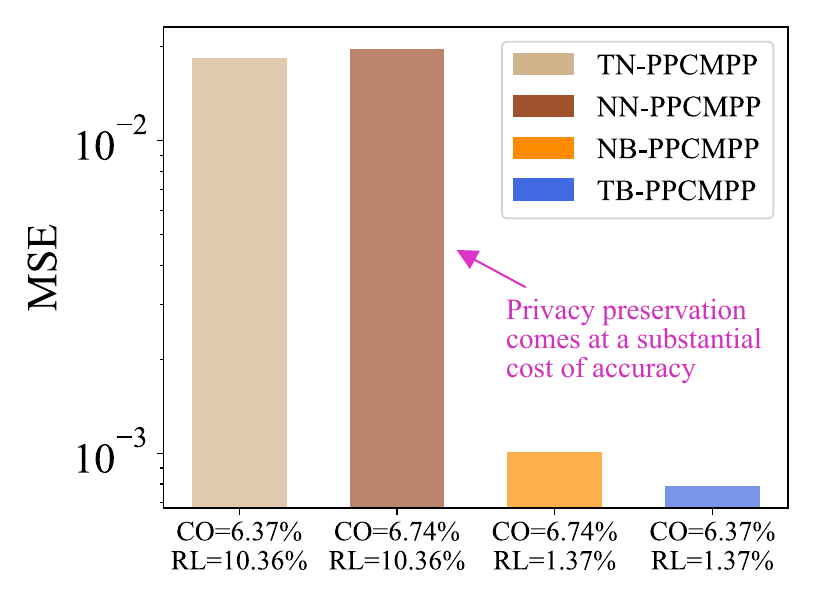}
\label{subfig2}}  
\setlength{\subfigcapskip}{-1.2em}
\subfigure[]{\hspace*{1em}\includegraphics[scale=0.38]{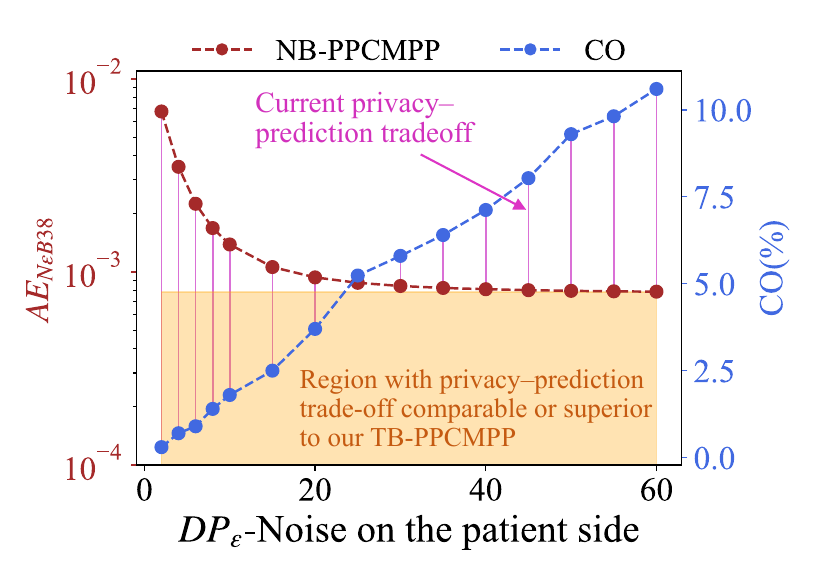}
\label{subfig: AE_CO_patientDP_text}} 
\setlength{\subfigcapskip}{-1.2em}
\subfigure[]{\includegraphics[scale=0.37]{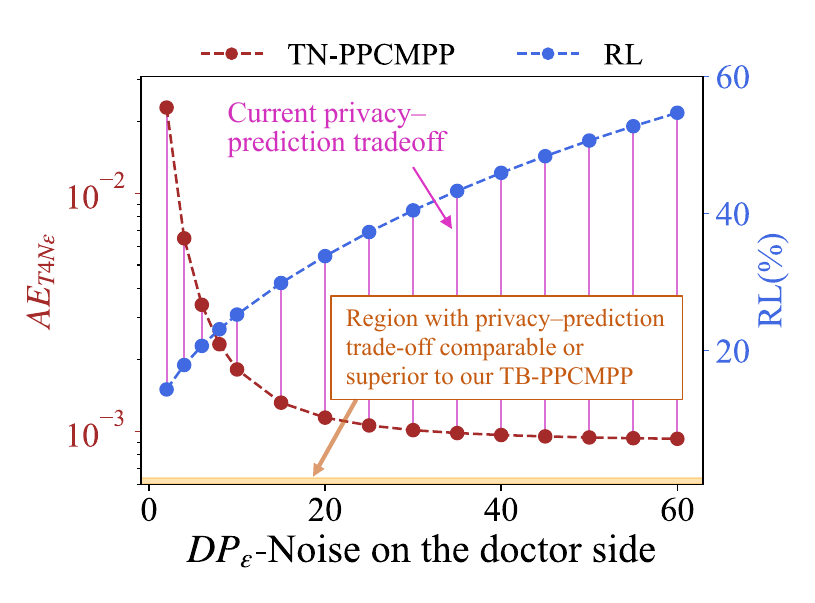}
\label{subfig: AE_RL_doctoruseDP_text}} 
\end{figure}
\vspace{-0.8cm} 

In the second experiment, we compare TN-PPCMPP, NB-PPCMPP, NN-PPCMPP, and TB-PPCMPP. As shown in Figure~\ref{fig:power_of_PPCMP}(d), where “N” denotes multiplicative noise, although the privacy level of TB-PPCMPP is controlled to be the highest, it still achieves the best prediction performance among the four approaches. Moreover, we observe that the other approaches, when adjusted to reach a similar level of privacy as TB-PPCMPP, suffer from significant losses in accuracy, underscoring the power of incorporating the TQMA-BSTD mechanism into PPCMPP.
Furthermore, we evaluate the privacy–prediction trade-off performance of NB-PPCMPP relative to the TB-PPCMPP when ``N'' refers to \textit{DP}$_{\epsilon}$-Noise. As shown in Figures~\ref{fig:power_of_PPCMP}(e) and \ref{fig:power_of_PPCMP}(f), the purple vertical lines represent the privacy–prediction trade-off achieved under specific $\epsilon$ values, while the orange shaded region marks where the trade-off is comparable to or better than that of the current TB-PPCMPP. We observe that the trade-offs under NB-PPCMPP and TN-PPCMPP both fail to fall within the orange region, indicating that achieving either an ideal level of privacy preservation or high prediction performance comes at the substantial cost of sacrificing the other. This highlights the advantage of TB-PPCMPP in balancing privacy and prediction.

\subsection{Real-World Data Analysis}\label{real data}
We explore the clinical implications of TB-PPCMPP on a real-world warfarin dataset  \citep{international2009estimation}. 
For comparison, we adopt five other methods: a model with a fixed dose of 35 $mg/week$, linear regression (LR) built on the entire dataset, NB-PPCMPP and TN-PPCMPP (where ``N'' refers to $DP_{\epsilon}$-Noise), and original CMPP. We control the privacy parameter such that TB-PPCMPP achieves the highest level of privacy preservation among the three privacy strategies to ensure a fair comparison.

\vspace{-0.3cm} 
\begin{figure}[H]
	\centering
	\caption{Results  on Warfarin Dataset.}
	\setlength{\subfigcapskip}{-1.2em}
	\subfigure{\includegraphics[scale=0.26]{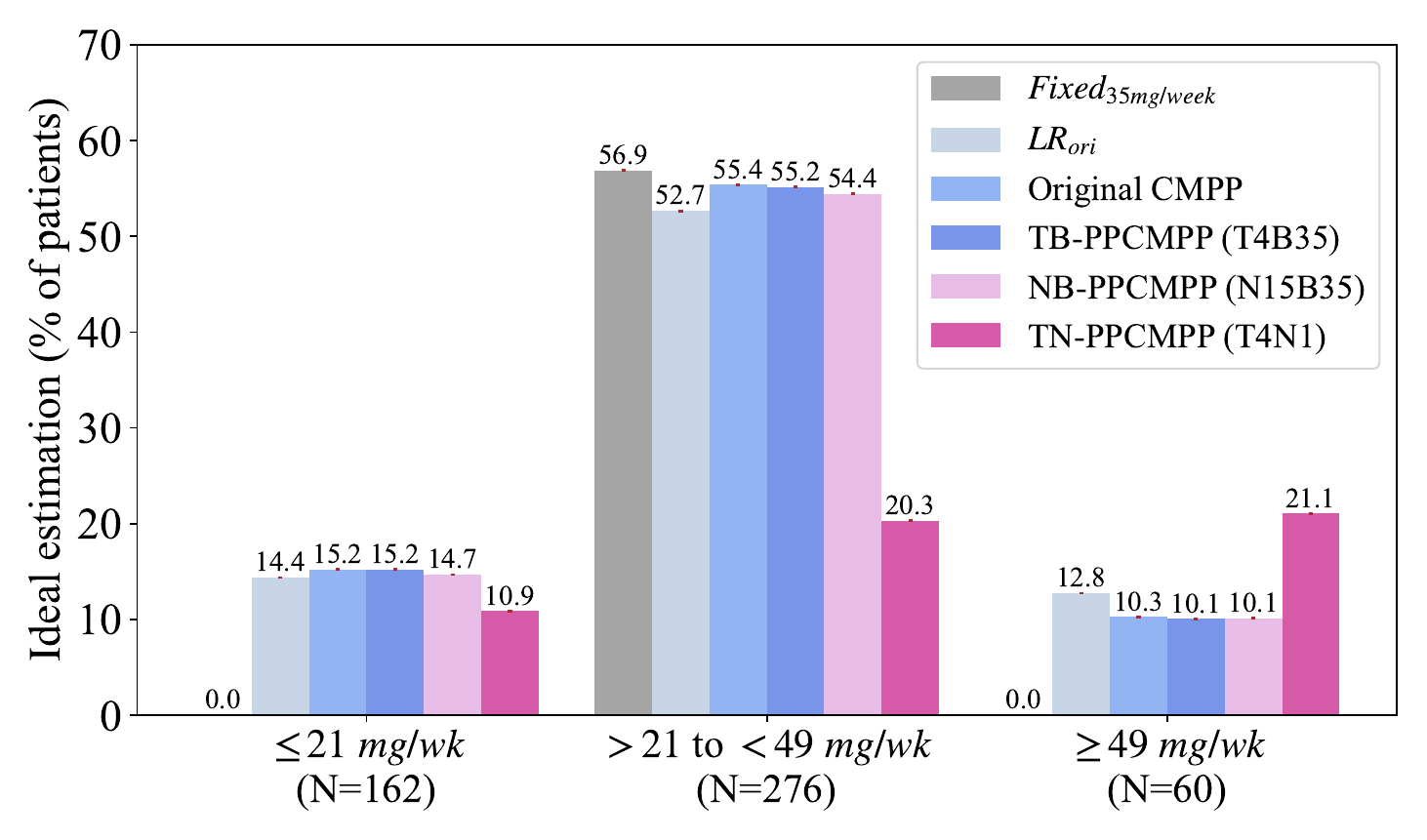}
		\label{subfig: clinical_1_utility}} 
	\setlength{\subfigcapskip}{-1.2em}
	\subfigure{\includegraphics[scale=0.32]{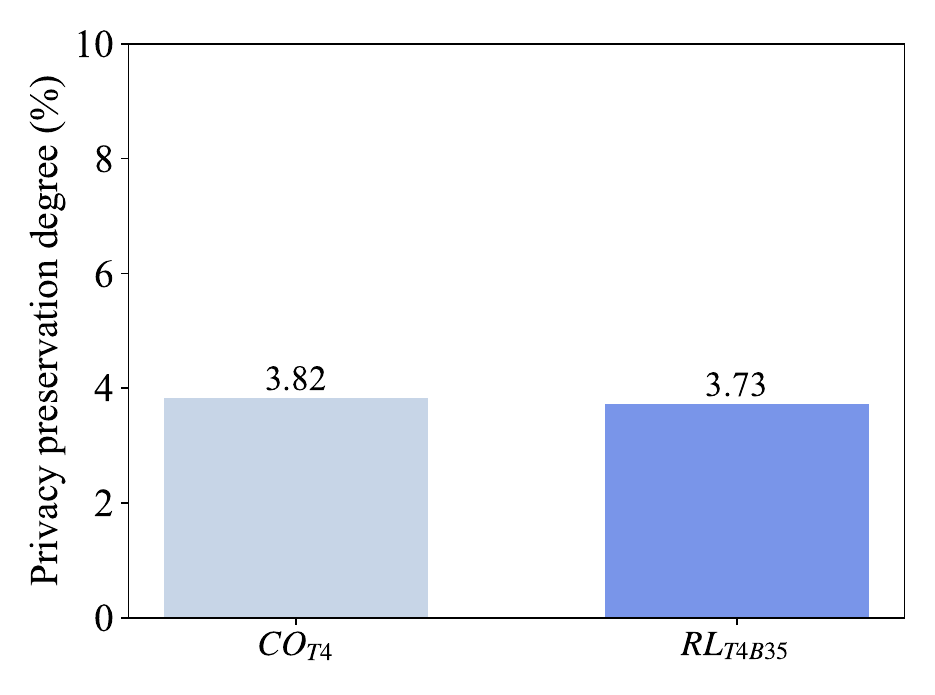}
		\label{subfig: clinical_1_privacy}}  
	\label{fig:clinical_1}
\end{figure}
\vspace{-0.8cm} 


Figure \ref{fig:clinical_1} shows the comparison, where $Fixed_\mathrm{35mg/week}$ corresponds to the fixed-dose model and $LR_\mathrm{ori}$ is the MSE of LR on the original dataset.
We find that: (1) For the fixed-dose model, none of the estimates for low- and high-dose groups are ideal, emphasizing the importance of developing a predict model.
(2) 
$AE_\mathrm{ori}$ outperforms the global result $LR_\mathrm{ori}$ in the low- and intermediate-dose groups and shows minimal difference in the high-dose group, highlighting the benefit of collaboration.
(3) With CO and RL as low as $3.82\%$ and $3.73\%$, respectively, the MSE of TB-PPCMPP is nearly identical to that of the original CMPP, indicating that TB-PPCMPP not only preserves privacy but also maintains accuracy.
(4) The ideal estimation of NB-PPCMPP and TN-PPCMPP is consistently inferior to that of TB-PPCMPP. Notably, the high ideal estimation observed for TN-PPCMPP in the high-dose group is not due to accurate predictions but rather to an upward bias introduced by the added noise. This result highlights the effectiveness of TB-PPCMPP in maintaining prediction performance.
We note that the generally inaccurate predictions observed here are primarily due to the lack of comprehensive consideration of demographic, medications taken, phenotypic, and genotype information.actors, medications taken, phenotypic characteristics, and genotype information.

\vspace{-0.4cm}
\begin{table}[H]
    \caption{Results of CMPP and PPCMPP}\label{Tab:Clinical_2}
    \renewcommand\arraystretch{1.1}
    \footnotesize
    \begin{center}
        \scalebox{0.91}{ 
        \begin{tabular}{lccc|ccc}
            \toprule
            \multirow{2}{*}{\textbf{Performance indicators}} & \multicolumn{3}{c}{\textbf{Low-dose group (size: 161)}} & \multicolumn{3}{c}{\textbf{High-dose group (size: 60)}} \\
            \cmidrule(lr){2-4} \cmidrule(lr){5-7}
            & \!\!{\makecell[l]{Original \\ CMPP}}\!\! & \!TB-PPCMPP\! & \!NN-PPCMPP\! & {\makecell[l]{Original \\ CMPP}}\!\! & \!TB-PPCMPP\! & \!NN-PPCMPP\! \\
            \midrule
            RMSE  & 0.0569 & \textbf{0.0567} & 0.0609 & 0.1259 & \textbf{0.1261} & 0.1194 \\
            No. of patients predicted to require extreme doses & 49 & \textbf{50} & 45 & 5 & \textbf{5} & 9 \\
            {\makecell[l]{No. of patients correctly predicted to require extreme doses}} & 38 & \textbf{38} & 34 & 3 & \textbf{3} & 5 \\
            {\makecell[l]{Per. of patients correctly predicted to require extreme doses}} & 23.46\% & 23.46\% & 21.19\%  & 5.00\% & 5.00\% & 7.60\%\\
            \bottomrule
        \end{tabular}}
        \vspace{3.5pt} 
        \footnotesize
        \begin{tabular}{@{}l@{}}
            \multicolumn{1}{@{}p{0.97\linewidth}@{}}{
                \scriptsize
                \textbf{Note}: NN-PPCMPP refers to the CMPP equipped with the \textit{Mul}$_{p_{noise}}$-Noise. We controlled NN-PPCMPP to achieve a CO of 3.97\%. Due to the current limitations on the number of doctors, RL could not be adjusted to match the level seen in TB-PPCMPP. Instead, we opted for a lower noise level with a variance of 0.03.
            }
        \end{tabular}
    \end{center}
\end{table}
\vspace{-0.85cm}


Table~\ref{Tab:Clinical_2} presents the results of CMPP, TB-PPCMPP, and NN-PPCMPP on two extreme groups: the low-dose group and the high-dose group. We ensured that both the CO and RL of TB-PPCMPP were lower than those of NN-PPCMPP. Even so, compared to NN-PPCMPP, TB-PPCMPP shows only minor differences from CMPP in the number of patients predicted to require extreme doses and achieves the same number of correctly predicted patients in both groups, demonstrating its potential for clinical application. Note that NN-PPCMPP performs better in the high-dose group, primarily because the added noise introduces an upward bias in the predictions.

\section{Conclusions and Extensions}\label{Conclusions}

Online collaborative medical prediction platforms are becoming increasingly popular in daily life due to their advantages, such as user-friendliness and real-time feedback.  
However, their further development is hindered by growing privacy concerns and limited prediction accuracy.  
This study designs a TQMA-BSTD mechanism and combines it with a delicate one-shot distributed learning framework, and then proposes a novel one-shot swapped distributed learning framework with input perturbation. 
Collaborative medical prediction platforms under the proposed distributed learning framework can defend against attribute attacks targeting patients and resist model extraction attacks targeting doctors, all without sacrificing prediction performance.

Our study also generates several opportunities for future research. First, our preservation mechanism is not limited to medical prediction or specific local algorithms; it can be applied to other online collaborative prediction platforms requiring high accuracy and strong privacy preservation.
Second, platforms could establish stricter qualification to determine the effective sample size contributed by each doctor participating in collaboration,  
as well as implement an accountability mechanism to identify and exclude doctors who are frequently dishonest or make incorrect decisions during the collaborative process.
Third, beyond the regression problems addressed in this paper, it is also crucial to develop preservation mechanisms for classification problems.

\bibliographystyle{informs2014}
\bibliography{ref_msom} 
\newpage



\begin{APPENDIX}{}

\section*{Appendix A: Algorithms}

This section first introduces the TQMA mechanism for defending against attribute attacks, followed by the BSTD mechanism for countering model extraction attacks. Finally, it presents the PPCMPP with the TQMA–BSTD preservation mechanism.

\vspace{-0.1in}
${\qquad}$
\hrule
\vspace{0.08in}
\textbf{Algorithm 1: TQMA}
\vspace{0.05in}
\hrule
\vspace{0.1in}
\textbf{Input}: A patient's query $x$ with the QIA value $v \in [a,b]$ for $a<b$, and the tree depth $k$\par
\textbf{1. Partition}: Split $[a, b]$  recursively, and obtain $2^k$ intervals:
\vspace{-0.1in}
$$
\{[a+(b-a)j2^{-k},a+(b-a)(j+1)2^{-k}]:j=0,\dots,2^{k}-1\}
\vspace{-0.1in}
$$
\quad	\textbf{2. Perturbation}: Perturb the value $v$ to the midpoint of its located interval $[a+(b-a)j 2^{-k},a+(b-a)(j+1)2^{-k}]$, where $0\leq j\leq 2^k-1$, that is, $v^{\text{TQMA}(k)}:=a+(b-a)\frac{2j+1}22^{-k}$.

\textbf{Output}: The perturbed query $x^{\text{TQMA}(k)}$ with the perturbed QIA value $v^{\text{TQMA}(k)}$
\vspace{0.2cm}
\hrule

\vspace{0.2in}

${\qquad}$
\hrule
\vspace{0.08in}
\textbf{Algorithm 2: BSTD mechanism}
\vspace{0.05in}
\hrule
\vspace{0.1in}
\textbf{Input}: The number of doctors $m$, a set of real numbers $\{a_1,\dots,a_m\}$, where $a_j$ is the prediction made by the $j$th doctor\par

\textbf{1. Threshold decryption}: 
Implement the $m$-out-of-$m$ threshold scheme; that is, the algorithm proceeds only when all doctors agree to start the collaboration.

\textbf{2. Presetting bounds}: Set the lower bound $p_\mathrm{lower}$  and  upper bound $p_\mathrm{upper}$ of bounded swapping, where $1 \leq p_\mathrm{lower}< p_\mathrm{upper}<m$.

\textbf{3. Ranking}: Rank $\{a_1,\dots,a_m\}$  to $\{a_1^*,\dots,a_m^*\}$, where $a_1^*\geq\dots\geq a_m^*$.

\textbf{4. Swapping}: For any $j\in \Lambda_0:=\{1,\dots,m\}$, define
\vspace{-0.1in}
$$
SW_j:=\{a^*_{j- p_\mathrm{upper}},\dots,a^*_{j-p_\mathrm{lower}},a^*_{j+p_\mathrm{lower}},\dots,a^*_{j+p_\mathrm{upper}}\} \cap\{a_1^*,\dots,a_m^*\}\backslash \{j\},
\vspace{-0.1in}
$$
randomly select $a_{k_j}^*\in SW_j$ without replacement, and swap $a_j^*$ with $a_{k_j}^*$.  Write $b_j=a_{k_j}^*$ and $b_{k_j}=a_j^*$. Iteratively define   $\Lambda_{2j-2}=\Lambda_{2j}\backslash \{j,k_j\}$.
Repeat the above swapping procedure until $SW_{j_{stop}}=\varnothing$ for some $j_{stop}\leq [m/2]$. If  $a_k^*$ remains after the swapping procedure, set $b_{k}=a_k^*$.

\textbf{Output}: The swapped set $\{b_1,\dots,b_m\}$ 
\hrule
${\qquad}$

${\qquad}$
\hrule
\vspace{0.08in}
\textbf{Algorithm 3: PPCMPP with TQMA–BSTD preservation mechanism}
\vspace{0.06in}
\hrule
\vspace{0.1in}
\textbf{Input}: A patient's query $x$, the tree depth $k$ of TQMA, bounded parameters $p_\mathrm{lower}$ and $p_\mathrm{upper}$ for BSTD, where $1 \leq p_\mathrm{lower}< p_\mathrm{upper}<m$.  \par

\textbf{Initialization}: Let $m$ be the number of doctors participating in CMPP who agree to adopt the current BSTD mechanism.  CMPP evaluates the qualification of the $j$th doctor to mimic their data size $|D_j|$ and sends both $|D_j|$ and $|D| = \sum_{j=1}^m |D_j|$ to the $j$th doctor.
\par

\textbf{1. Privacy preservation for patients}: The platform sends the TQMA perturbed input $x^{TQMA(k)}$  to $m$ participating doctors.\par


\textbf{2. Local processing}: The $j$th doctor autonomously determines the learning algorithm,
autonomously trains the local parameter ${h}_j$ and refines it to
$\hat{h}_j=({h}_j)^{\log_{|D_j|}|D|}$. 
Then, the $j$th doctor deduces a local estimator $f_{D_j,\hat{h}_j}(x^{TQMA(k)})$.
\par

\textbf{3. Privacy preservation for doctors}: 
Doctors submit the $f_{B_j}(x^{\text{TQMA}(k)})$ (i.e., $\frac{|D_j|}{|D|}f_{D_j,\hat{h}_j}(x^{\text{TQMA}(k)})$)
 to the BSTD mechanism,  which then transforms this value into the swapped version  $f_{B_j}^{p_\mathrm{lower},p_\mathrm{upper}}(x^{\text{TQMA}(k)})$.
\par

\textbf{4. Communication and qualification}: The BSTD mechanism transmits all swapped outputs $\{f_{B_j}^{p_\mathrm{lower},p_\mathrm{upper}}(x^{\text{TQMA}(k)})\}_{j=1}^m$ to the central agent. The central agent labels the  $j$th doctor as ``active'' if $|f_{B_j}^{p_\mathrm{lower},p_\mathrm{upper}}(x^{\text{TQMA}(k)})|\geq \frac{|D_j|}{|D|^2}$ and rearranges all active doctors as $\{1,\dots,m^*\}$ with data $\{D_1^*,\dots,D_{m^*}^*\}$.\par

\textbf{5. Synthesis}: The central  agent   synthesizes all  active local outputs as 
\begin{equation}\label{Global-estimator-*}
\overline{f}_{D}(x_i)=\sum_{j=1}^{m^*}\frac{|D|f_{B_j}^{p_{lower},p_{upper}}(x^{TQMA(k)})}{|D^*|},
\end{equation}
where   $D^*=\bigcup_{j=1}^{m^*}D^*_j$.\par
\textbf{ Output}: The synthesized estimator $\overline{f}_{D}(x_i)$
\vspace{0.2cm} 
\hrule
${\qquad}$

\section*{Appendix B: Experimental Settings}\label{Experimental_Settings}
This section describes the experimental settings for both the toy simulations and the real-world data analysis.
In all experiments, prediction accuracy is evaluated using the mean squared error (MSE), defined as $\frac{1}{N'} \sum_{i=1}^{N'} (y_{i} - \overline{f}_{D}(x_{i}))^{2}$. The average error (AE) refers to the MSE obtained from each corresponding algorithm.
Each experiment is repeated 20 times to compute average results, and the parameters of the learning algorithms are trained using five-fold cross-validation. All experiments are conducted using Python 3.7 on a PC equipped with an Intel Core i5 2 GHz processor.

We assume that each doctor holds a different data size. Given the total number of samples $|D|$ and the number of doctors $m$, we randomly select $|D_1|, \cdots, |D_{m-1}|$ from the range $\left[\frac{0.8|D|}{m}, \frac{|D|}{m}\right]$ following a uniform distribution, and set $|D_m| = |D| - \sum_{j=1}^{m-1} |D_j|$ to reflect the autonomy of individual doctors.  
Specifically, we assume that the central agent targets one doctor, whose data size is 1,322, for model extraction attacks.  
To simulate the decision-making process of the $j$th doctor, we randomly select a local algorithm from Table~\ref{Tab:application1}. Note that only the attack and test samples are used for perturbation.

\subsection*{B.1: Experimental Settings of Toy Simulations}

This section presents the attribute and model extraction attacks simulated in the toy experiments.

\begin{itemize}
	\item \textit{Simulate $\mu$-attribute attacks.} 
	We simulate different QIA values held by attackers by adding Gaussian noise $\mathcal{N}(0,\sigma^2)$ with $\sigma= 10^{-3}$ to patients' QIA values. Attackers conduct $\mu$-attribute attacks with their preference of $\mu$. CO measures the likelihood of patients experiencing these attacks.
	
	\item \textit{Simulate $\varepsilon$-model extraction attacks.} The central agent targets the $j$th doctor
	and prepares $|D_j|$ fake queries to obtain input–output pairs to build a model using NWK (Gaussian) that approximates this doctor's model.
	The central agent then replaces this doctor in CMPP with the input–output pairs as the local dataset. RL measures the likelihood of finding the correct input–output correspondence.
\end{itemize}

\subsection*{B.2: Experimental Settings of Real-world Data Analysis}

This section describes the experimental setup on the warfarin dataset.
The dataset contains 5,700 medical records, which are artificially distributed across 10 local agents to simulate a collaborative medical prediction scenario.
In the experiments, we consider variables including age, height, weight, race, medications taken, and therapeutic dose.
We remove one outlier with an extraordinarily high dose of 315~mg/week, convert age intervals to their corresponding medians, and transform two nominal attributes into numerical form.
Subsequently, we normalize the data and randomly split it into three parts: approximately 77\% for training, 14\% for attack scenarios, and 9\% for testing.
This division is repeated 20 times to obtain averaged results.

We divide the testing samples into three groups based on the actual required dose: low-dose group ($\leq$21 $mg/week$), intermediate-dose group ($>$21 and $<$49 $mg/week$), and high-dose group ($\geq$49 $mg/week$). We assess the prediction of above models on each group by calculating the percentages of ideal estimation (within $20\%$ of the actual dose), underestimation (at least 20$\%$ lower than the actual dose), and overestimation (at least 20$\%$ higher than the actual dose). We use the value 20\% because it represents 
a difference clinicians would be likely to define as clinically relevant.

\section*{Appendix C: Additional Experimental Results}

\subsection*{C.1: Selection of Privacy Parameters for TB-PPCMPP}

This simulation, together with the results in Figure~\ref{fig:power_of_PPCMP}, illustrates how to determine the privacy parameters in TB-PPCMPP.  

For the selection of the tree depth $k$, based on Figure~\ref{fig:power_of_PPCMP}(a) and Figure~\ref{fig:parameter_TB_PPCMPP}(a), we see that as $k$ increases, $AE_\mathrm{Tk}$ quickly approaches  $AE_\mathrm{ori}$  and then barely improves while CO continues to deteriorate, which verifies our theoretical findings in Theorem \ref{Theorem:Optimal} that when $k\geq \frac{\log_2|D|}{4r+2d}-1$, as $k$ increases, the prediction on perturbed data has the same performance as on original data while CO gradually becomes larger. 
 We set $k$ to 4 since the high prediction accuracy and high preservation level of patients coexist at this point.
 
For the selection of $p_\mathrm{lower}$ and $p_\mathrm{upper}$, Figure~Figure~\ref{fig:parameter_TB_PPCMPP}(b) and Figure~\ref{fig:parameter_TB_PPCMPP}(c)  show that $AE_\mathrm{T4B}$ remains nearly unchanged,  
while $RL$ drops sharply as $p_\mathrm{lower}$ increases from 1 to 2 and consistently stays below $1.70\%$ when $p_\mathrm{lower}$ is set to 3.  
We finally set $p_\mathrm{lower}$ to 3 and $p_\mathrm{upper}$ to 8 to introduce more randomness to avoid model extraction attacks.

\vspace{-0.2cm} 
\begin{figure}[H]
	\centering
	\caption{Determining tree depth $k$, swapping bounds $p_\mathrm{lower}$ and $p_\mathrm{upper}$ for TB-PPCMPP}	\label{fig:parameter_TB_PPCMPP}
	\setlength{\subfigcapskip}{-1.2em}
	\subfigure[Role of tree depth $k$]{\includegraphics[scale=0.30]{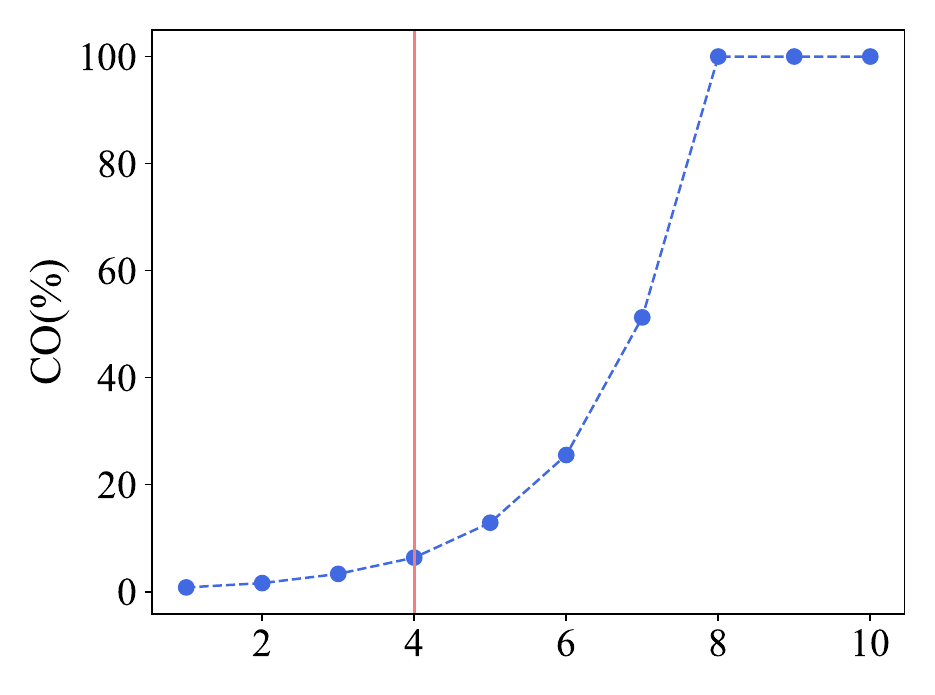}
		\label{subfig: TQMA_k_2}} 
	\setlength{\subfigcapskip}{-1.2em}
	\subfigure[Role of $p_\mathrm{lower}$ ($p_\mathrm{upper} = p_\mathrm{lower}+1$)]{\includegraphics[scale=0.31]{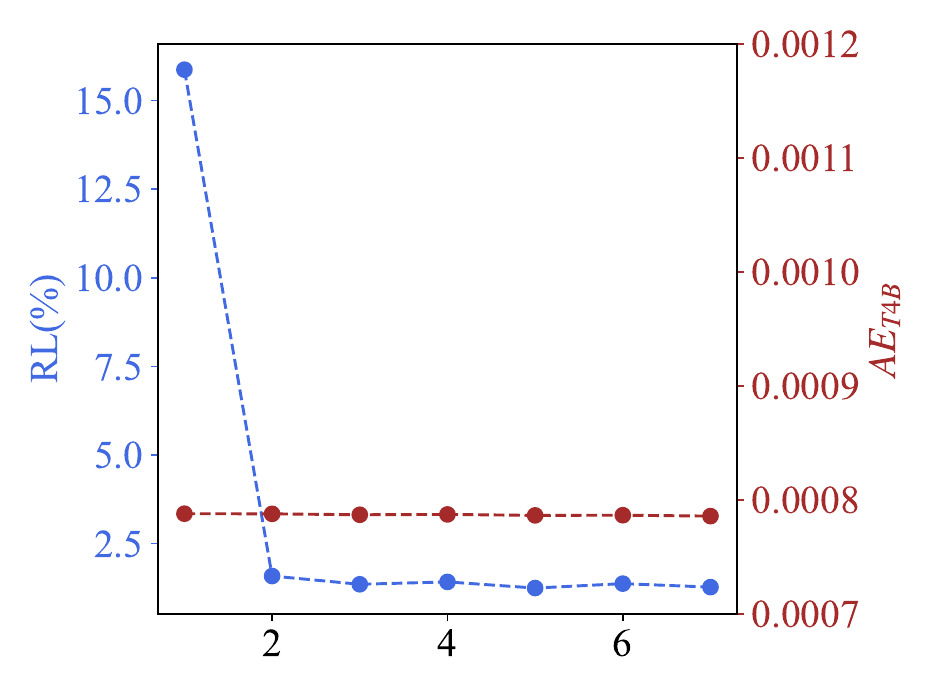}\label{subfig: RL_dis1_AE_2}} 
	\setlength{\subfigcapskip}{-1.2em}
	\subfigure[Role of $p_\mathrm{upper}$ ($p_\mathrm{lower}=3$)]{\includegraphics[scale=0.31]{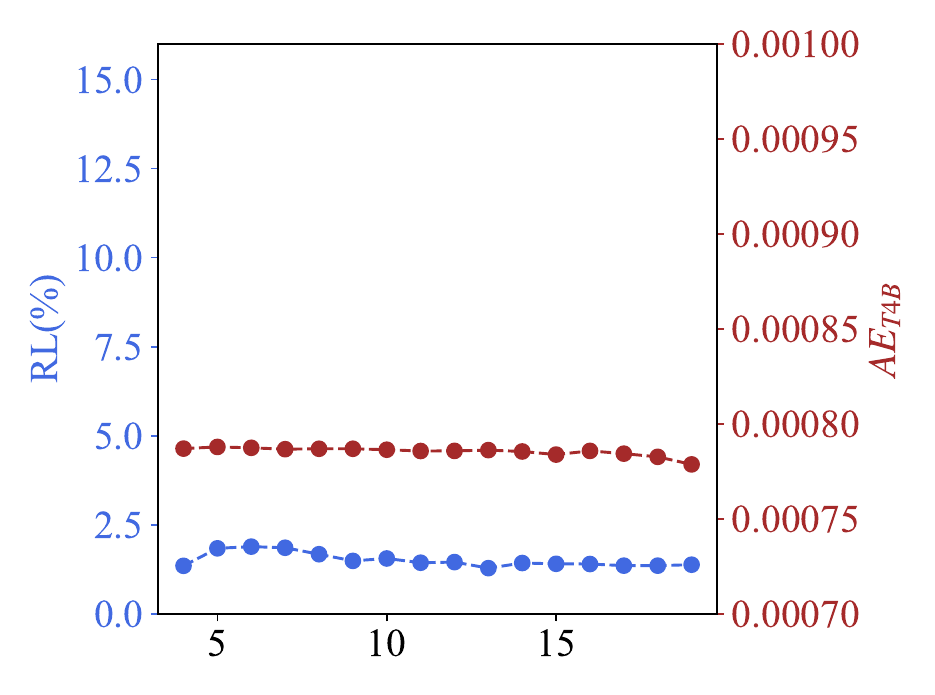}\label{subfig: RL_lower3_AE}}  
	\vspace{-0.2cm}
\end{figure}
\vspace{-0.3cm}

\subsection*{C.2: Effectiveness of TB-PPCMPP on the Other Tree-based Learning Algorithm}\label{regression_tree}

This simulation evaluates the performance when using a regression tree as the local algorithm to demonstrate the generalizability of the proposed TQMA–BSTD mechanism.  
As shown in Figure~\ref{fig:effective_CMPT_regression_tree}, $AE_\mathrm{T4B38}$ is nearly identical to $AE_\mathrm{ori}$, with only a 0.33\% change.  
The CO remains unchanged compared to the previous results, while RL decreases to 1.73\%.  
These results demonstrate the effectiveness of TB-PPCMPP in preserving privacy without sacrificing accuracy and highlight the generalizability of the TQMA–BSTD mechanism, as it imposes no restrictions on the choice of local algorithms.

\vspace{-0.3cm} 
\begin{figure}[H]
	\centering
	\caption{Effectiveness of TB-PPCMPP (Assume the local algorithm in CMPP is regression tree).}
	\setlength{\subfigcapskip}{-1.2em}
	\subfigure{\includegraphics[scale=0.31]{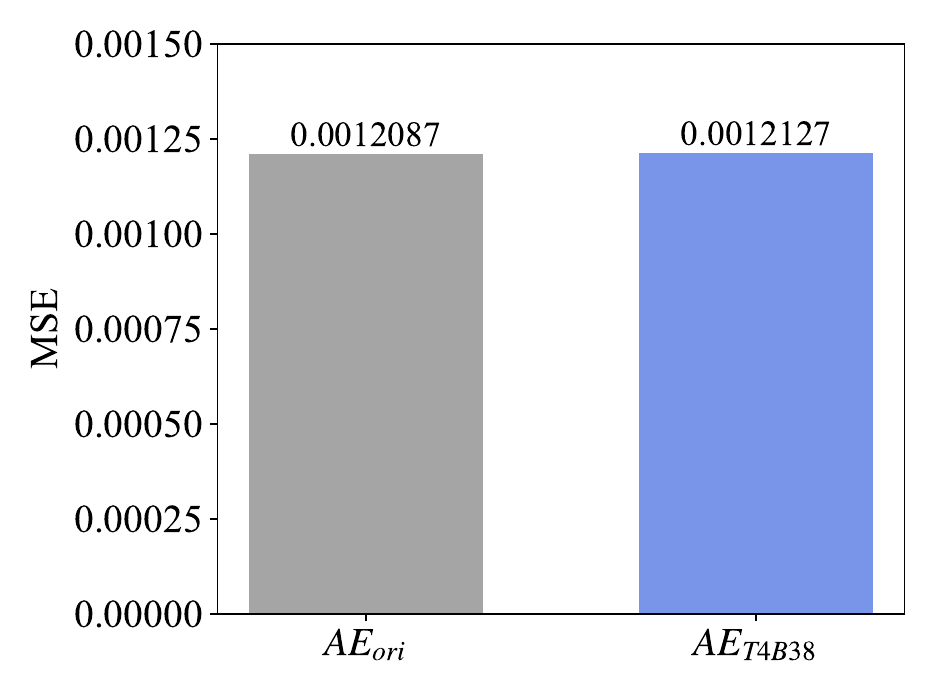}
		\label{subfig: CMPT_MSE_regression_tree}} 
	\setlength{\subfigcapskip}{-1.2em}
	\subfigure{\includegraphics[scale=0.31]{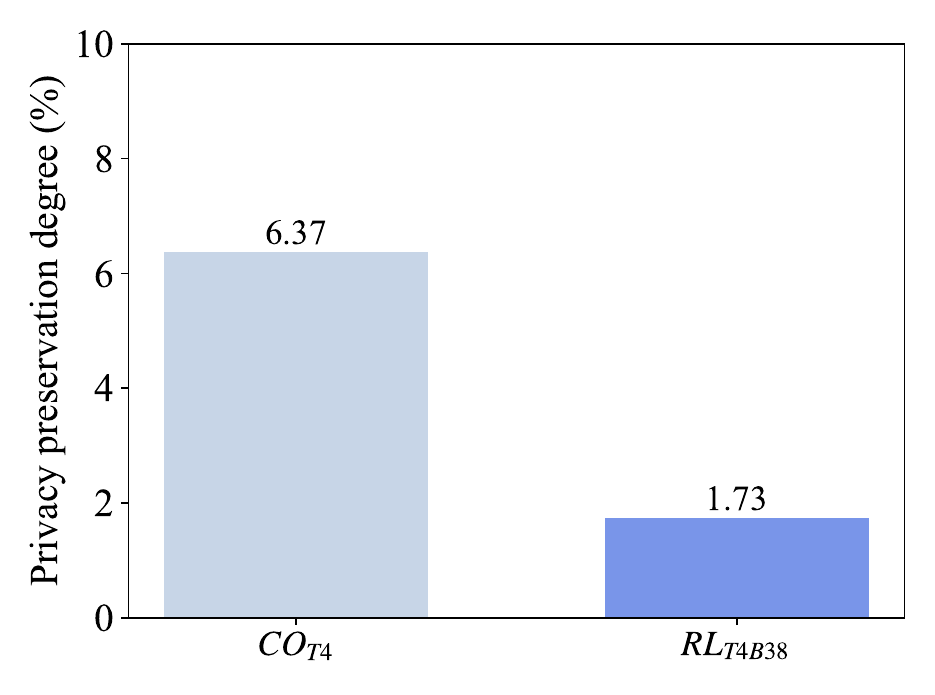}
		\label{subfig: CMPT_CO_RL_regression_tree}}  
	\label{fig:effective_CMPT_regression_tree}
\end{figure}
\vspace{-0.8cm}

\subsection*{C.3: Additional Notes on the Results in Table \ref{tab:attack_comparison}}
We finally provide an explanation for an interesting phenomenon observed in Table~\ref{tab:attack_comparison}: after applying \textit{DP}$_{\epsilon}$-Noise, the prediction performance under model extraction attacks was unexpectedly better than without the attack (see the $\star$-marked results). This is because, in our setup, the doctor targeted by the attack holds a relatively larger amount of data, and using this doctor’s model alone can sometimes outperform distributed learning. Essentially, after the attack, the substituted model can be regarded as one trained on noise-free data, and this effect becomes more pronounced as the noise level increases.

\section*{Appendix D: Proofs of Theoretical Results}

In this appendix, we devote to proving the theoretical results. To this end, some preliminaries are need. Given a $\tau\geq 0$, the $j$th doctor is ``$\tau$-active'' if $|f_{D_j, h_j}(x)|\geq \tau$.
Rearrange all the $\tau$-active doctors as $\{1,\dots,m^\diamond\}$ with corresponding dataset $\{D_1^\diamond,\dots,D_{m^\diamond}^\diamond\}$.
Define
\begin{equation}\label{global-without-privacy-mu}
\tilde{f}_{D,\tau}(x)=\sum_{j=1}^{m^\diamond}\frac{|D_j^\diamond|}{|D^\diamond|}f_{D_j,{h}_j}(x),
\end{equation}
where $D^\diamond=\bigcup_{j=1}^{m^\diamond}D_j^\diamond$. 
Then it can be derived from \eqref{global-without-privacy-mu} and \eqref{local-estimation} that 
$$
\tilde{f}_{D,\tau}(x)=\sum_{j=1}^m\sum_{i=1}^{|D_j|}\frac{I_{|f_{D_j,h_j}(x)|\geq \tau}}{\sum_{j=1}^mI_{|f_{D_j,h_j}(x)|\geq \tau}} W_{j,h_j,x_{i,j}}(x)y_{i,j}.
$$
Writing
\begin{equation}\label{def.big.w}
\mathcal W_{\tau,D_j, {h}_j,x_{i,j}}(x):= \frac{I_{|f_{D_j,h_j}(x)|\geq \tau}}{\sum_{j=1}^mI_{|f_{D_j,h_j}(x)|\geq \tau}}W_{j,h_j,x_{i,j}}(x),
\end{equation}
we have
\begin{equation}\label{rewrite-f}
\tilde{f}_{D,\tau}(x)=\sum_{j=1}^m\sum_{i=1}^{|D_j|}\mathcal W_{\tau,D_j, {h}_j,x_{i,j}}(x)y_{i,j}.
\end{equation}
In this way, $\tilde{f}_{D,\tau}(x)$ can be regarded as a new local average regression estimate for the  sample $D$. Therefore, the local agents that are not activated also play  important roles in determining the position information of $x$. We then deduce some
important properties of the weight $\mathcal W_{\tau,D_j, {h}_j,x_{i,j}}(x)$. For this purpose, we should  introduce some important  properties of $ W_{j,h_j,x'}(x)$ given in Table \ref{Tab:application1}. 

(A) There exists a univariate decreasing function $\xi_j(\cdot):\mathbb R_+\rightarrow \mathbb R_+$ such that
\begin{equation}\label{weight:localization}
W_{j,h_j,x'}(x)\leq \xi_j, \qquad \mbox{for} \  \|x-x'\|\geq \tilde{c}_j h_j,
\end{equation}
where $\|\cdot\|$ denotes the Euclidean norm on $\mathbb R^d$ and $\tilde{c}_j$ is a constant depending only on $d$.

(B) Let $B_{h_j}(x)$ be the Euclidean ball with center $x$ and radius $h_j$, i.e., $B_{h_j}(x):=\{x':\|x'-x\|\leq h_j\}$ and $\Lambda_j:=\{x:(x,y)\in D_j\}$. If $B_{\tilde{c}_j h_j}(x)\cap \Lambda_j\neq\varnothing$, then
\begin{equation}\label{weight:summation}
\sum_{i=1}^{|D_j|}W_{j,h_j,x_{i,j}}(x)=1.
\end{equation}

(C) For any $x\in\mathbb I^d$, there are absolute constants $\bar{c}_j$ and $\bar{c}_j'$ such that $0\leq W_{j, h_j, x_{i,j}}(x)\leq 1$ and
\begin{equation}\label{weight:mass}
\sum_{i=1}^{|D_j|}W^2_{j,h_j,x_{i,j}}(x)\leq \frac{\bar{c}_jI_{|\tilde{A}_{j}(x)\cap\Lambda_j|\neq 0}}{|\tilde{A}_{j}(x)\cap\Lambda_j|},
\end{equation}
where $I_A$ denotes the indicator on the event $A$, $0/0:=0$ and $\tilde{A}_{j}(x)\ni x$ is a compact subset of $\mathbb I^d$ with volume $\tilde{c_j}'h_j^d.$

With these helps, we are in a position to present the property of $\mathcal W_{\tau,D_j, {h}_j,x_{i,j}}(x)$ defined in \eqref{def.big.w}.

\begin{lemma}\label{Lemma:new-weights}
	If $W_{j,h_j,x_{i,j}}(x)$ is given in Table \ref{Tab:application1}, then  we have
	\begin{equation}\label{weight:localization-big}
	\sum_{j=1}^m\sum_{i=1}^{|D_j|}\mathcal W_{\tau,D_j, {h}_j,x_{i,j}}(x)I_{\|x-x_{i,j}\|\geq \tilde{c}_jh_j} 
	\leq\max_{1\leq j\leq m}|D_j|\xi_j.
	\end{equation}
	Furthermore,
	if   there exists a $j$ such that $|f_{D_j,h_j}(x)|\geq \tau$ with $\tau>|D|^{-1}h_j^{2r}$,  then
	\begin{equation}\label{weight:summation-big}
	\sum_{j=1}^m\sum_{i=1}^{|D_j|}\mathcal W_{\tau,D_j,h_j,x_{i,j}}(x)=1.
	\end{equation}
\end{lemma}

{\bf Proof.} For $\|x-x'\|\geq \tilde{c}_j h_j$, it follows from (\ref{weight:localization}) that
$$
W_{j,h_j,x'}(x)\leq \xi_j.
$$
We then get from   (\ref{def.big.w})  that
$$
\mathcal W_{\tau,D_j, {h}_j,x_{i,j}}(x)\leq \frac{I_{|f_{D_j,h_j}(x)|\geq \tau}}{\sum_{j=1}^mI_{|f_{D_j,h_j}(x)|\geq \tau}}\xi_j,
$$
which implies
$$
\sum_{j=1}^m\sum_{i=1}^{|D_j|}\mathcal W_{\tau,D_j, {h}_j,x_{i,j}}(x)I_{\|x-x_{i,j}\|\geq \tilde{c}_jh_j}
\leq
\max_{1\leq j\leq m}|D_j|\xi_j.
$$
We then turn to proving (\ref{weight:summation-big}). If $B_{\tilde{c}_jh_j}(x)\cap \Lambda_j=\varnothing$ for all $j=1,\dots,m$, we have from (\ref{local-estimation}) and $\xi_j\leq h_j^{2r}(|D||D_j|M)^{-1}$ that
\begin{equation}\label{p.2.useful}
|f_{D_j,h_j}(x)|\leq\sum_{i=1}^{|D_j|}W_{j,h_j,x_{i,j}}(x)|y_{i,j}|
\leq|D_j|\xi_jM\leq h_j^{2r}/|D|,
\end{equation}
which contradicts the assumption    $|f_{D_j,h_j}(x)|\geq \tau>h_j^{2r}/|D|$. Therefore, there exists a $j$ such that  $B_{\tilde{c}h_j}(x)\cap \Lambda_j\neq\varnothing$.
This  implies \eqref{weight:summation}, which together with
(\ref{def.big.w})  yields
$$
\sum_{j=1}^m\sum_{i=1}^{|D_j|}\mathcal W_{\tau,h_j,D_j,x_{i,j}}(x)=
\sum_{j=1}^m\frac{I_{|f_{D_j,h_j}(x)|\geq \tau}}{\sum_{j=1}^mI_{|f_{D_j,h_j}(x)|\geq \tau}}\sum_{i=1}^{|D_j|}W_{j,h_j,x_{i,j}}(x)=1.
$$
This completes the proof of Lemma \ref{Lemma:new-weights}. $\Box$

We then present the following lemmas to ease our proofs. The first one can be found in \citep{liu2022enabling}.

\begin{lemma}\label{Lemma:important}
	For any  $\mathcal A\subseteq\mathbb I^d$ and $u\in\mathbb N_0:=\mathbb N\cup\{0\}$, there holds
	$$
	E\left[ \frac{ I_{|\mathcal A\cap\Lambda_j|\neq 0}}{ |\mathcal A\cap\Lambda_j|^u}\right]\leq      \frac{(u+1)!|D_j|!}{(|D_j|+u)!\left(\rho_X (\mathcal A)\right)^u}.
	$$
\end{lemma}

The second one is based on the standard statistical argument.

\begin{lemma}\label{lemma:control-force-1}
	For any $v\in\mathbb N$  and $x\in\mathbb I^d$, under   Assumption \ref{Assumption:distribution}, if $\tau>h_j^{2r}/{|D|}$ and $f_{D_j,h_j}$ is defined by \eqref{local-estimation} with $W_{j,h_j,x_{i,j}}(x)$ being given in Table \ref{Tab:application1}, then  for any $j=1,\dots,m,$ there holds
	\begin{equation}\label{lemma.2.1}
	\sum_{j=1}^mE\left[ \frac{ I_{|f_{D_j,h_j}(x)|\geq \tau}}{\left(\sum_{j=1}^mI_{|f_{D_j,h_j}(x)|\geq 1/\tau}\right)^v}\right]
	\leq \tilde{C}_1 \frac{v!m!}{(m+v-1)!}\frac{ 1}{  \min_{1\leq j\leq m} |D_j|^{v-1}(\tilde{c}_jh_j)^{(v-1)d}},
	\end{equation}
	where $\tilde{C}_1=\left(\Gamma(1+d/2)/(e\rho_{\min}\pi^{d/2})\right)^{v-1}$.
\end{lemma}
{\bf Proof.}
Denote $\Lambda_j:=\{x_{i,j}\}_{i=1}^{|D_j|}$.
Since $0/0=0$ in our definition, we have
\begin{align}
	&\sum_{j=1}^mE\left[ \frac{I_{|f_{D_j,h_j}(x)|\geq \tau}}{\left(\sum_{j=1}^mI_{|f_{D_j,h_j}(x)|\geq \tau}\right)^v}\right]
	=
	E\left[\frac{\sum_{j=1}^mI_{|f_{D_j,h_j}(x)|\geq \tau}}{\left(\sum_{j=1}^mI_{|f_{D_j,h_j}(x)|\geq \tau}\right)^v}\right]\notag\\
	&=E\left[\frac{1}{\left(\sum_{j=1}^mI_{|f_{D_j,h_j}(x)|\geq \tau}\right)^{v-1}}I_{\sum_{j=1}^mI_{|f_{D_j,h_j}(x)|\geq \tau}>0}\right] \notag\\
	&=\sum_{\ell=1}^mE\left[\frac{1}{\left(\sum_{j=1}^mI_{|f_{D_j,h_j}(x)|\geq  \tau}\right)^{v-1}}\big | \sum_{j=1}^mI_{|f_{D_j,h_j}(x)|\geq \tau}=\ell\right]
	P\left[\sum_{j=1}^mI_{|f_{D_j,h_j}(x)|\geq \tau}=\ell\right]\notag\\
	&=
	\sum_{\ell=1}^m \frac{1}{\ell^{v-1}}
	P\left[\sum_{j=1}^mI_{|f_{D_j,h_j}(x)|\geq \tau}=\ell\right]\notag.
\end{align}


Since $\tau>h_j^{2r}/{|D|}$, similar argument as that after  \eqref{p.2.useful} shows that 
$|f_{D_j,h_j}(x)|\geq \tau$ implies 
$B_{\tilde{c}_jh_j}(x)\cap\Lambda_j\neq\varnothing$.
Writing $\delta_j:=1-(1-\rho(\tilde{c}_jB_{h_j}(x)))^{|D_j|}$, we have
\begin{align}
	 P\left[\sum_{j=1}^mI_{|f_{D_j,h_j}(x)|\geq \tau}=\ell\right]
	\leq P\left[\sum_{j=1}^mI_{B_{\tilde{c}_jh_j}(x)\cap \Lambda_j\neq\varnothing}=\ell\right]\notag 
	 \leq
	\max_{1\leq j\leq m}\left(\begin{array}{c}
		m\\
		\ell
	\end{array}
	\right)
	\delta_j^\ell(1-\delta_j)^{m-\ell}\notag.
\end{align}
Therefore, we obtain
\begin{align}\label{p.2.1}
&\sum_{j=1}^mE\left[ \frac{I_{|f_{D_j,h_j}(x)|\geq \tau}}{\left(\sum_{j=1}^mI_{|f_{D_j,h_j}(x)|\geq \tau}\right)^v}\right]
\leq 
\max_{1\leq j\leq m}\sum_{\ell=1}^m \frac{1}{\ell^{v-1}}\left(\begin{array}{c}
m  \\
\ell
\end{array}
\right)
\delta_j^\ell(1-\delta_j)^{m-\ell}\nonumber\\
&\leq
\max_{1\leq j\leq m}\sum_{\ell=1}^{m} \frac{v!}{(\ell+1)\cdots(\ell+v-1)} \left(\begin{array}{c}{m} \\
\ell\end{array}\right) \delta_j^\ell
(1-\delta_j)^{m-\ell}\nonumber\\
&= 
\max_{1\leq j\leq m}\frac{v!m!}{(m+v-1)!\delta_j^{v-1}}\sum_{\ell=1}^{m}
\left(\begin{array}{c}{m}+v-1\\
\ell+v-1\end{array}\right)\delta_j^{\ell+v-1}(1-\delta_j)^{m-\ell}\nonumber\\
&\leq
\frac{v!m!}{(m+v-1)!\min_{1\leq j\leq m}\delta_j^{v-1}}.
\end{align}
Due to Assumption \ref{Assumption:distribution}, we have
$$
\frac{\rho_{\min}\pi^{d/2}}{\Gamma(1+d/2)} (\tilde{c}_jh_j)^d\leq  \rho(B_{\tilde{c}_jh_j}(x))\leq  \frac{ \rho_{\max}\pi^{d/2}}{\Gamma(1+d/2)} (\tilde{c}_jh_j)^d.
$$
Then
$$
\delta_j=1-(1-\rho(B_{\tilde{c}_jh_j}(x)))^{|D_j|}\geq
1-\left(\left(1-\frac{ \rho_{\min}\pi^{d/2}}{\Gamma(1+d/2)} (\tilde{c}_jh_j)^d\right)^{|D_j|}\right).
$$
Noting that $(1-a)^n\leq\frac{1}{ean}$ for $0< a\leq 1$, we obtain
$$
\delta_j\geq  1-\left(\left(1-\frac{ \rho_{\min}\pi^{d/2}}{\Gamma(1+d/2)} (\tilde{c}_jh_j)^d \right)^{|D_j|}\right)\leq
\frac{\Gamma(1+d/2)}{e\rho_{\min}\pi^{d/2} |D_j|(\tilde{c}_jh_j)^d}.
$$
Plugging the above estimate into (\ref{p.2.1}), we obtain (\ref{lemma.2.1}) and prove Lemma 
\ref{lemma:control-force-1}. $\Box$

Our third lemma focuses on the expectation of weight $\mathcal W^2_{\tau,D_j,h_j,x_{i,j}}(x)$.

\begin{lemma}\label{Lemma:expectation-weight}
	If $W_{j,h_j,x_{i,j}}(x)$ is given in Table \ref{Tab:application1}, then,
	$$
	E\left[\sum_{j=1}^m\sum_{i=1}^{|D_j|}\mathcal W^2_{\tau,D_j,h_j,x_{i,j}}(x)\right]
	\leq
	\tilde{C}_2\max_{1\leq j\leq m}\frac{1}{m|D_j|h_j^d},
	$$
	where $\tilde{C}_2:=6\tilde{C}_1^{1/2}/(\bar{c}_j'p_{\min}).$
\end{lemma}

{\bf Proof.}
It follows from (\ref{def.big.w}), H\"{o}lder inequality,  Lemma \ref{Lemma:important} with $\mathcal A= \tilde{A}_{j}(x)$ and $u=2$, Lemma \ref{lemma:control-force-1} with $v=3$ and \eqref{weight:mass} that
\begin{align}
	&E\left[\sum_{j=1}^m\sum_{i=1}^{|D_j|} \mathcal W^2_{\tau,D_j,h_j,x_{i,j}}(x)\right]
	=E\left[\sum_{j=1}^m\sum_{i=1}^{|D_j|}\frac{I_{|f_{D_j,h_j}(x)|\geq \tau}}{\left(\sum_{j=1}^mI_{|f_{D_j,h_j}(x)|\geq \tau}\right)^2} W^2_{j,h_j,x_{i,j}}(x)\right]\notag\\
	&\leq
	\left( E\left[\sum_{j=1}^m\frac{I_{|f_{D_j,h_j}(x)|\geq \tau}}{\left(\sum_{j=1}^mI_{|f_{D_j,h_j}(x)|\geq \tau}\right)^2}\right]^2\right)^{1/2}
	\max_{1\leq j\leq m}\left( E\left[\left(\sum_{i=1}^{|D_j|}W^2_{j,h_j,x_{i,j}}(x)\right)^2\right]\right)^{1/2}\notag\\
	&\leq
	\left(\tilde{C}_1   \frac{6}{m^2}\frac{ 1}{  \min_{1\leq j\leq m} |D_j|^{2}h_j^{2d}}\right)^{1/2}
	\max_{1\leq j\leq m}\left(\frac{6 }{ |D_j|^2 \left(\bar{c}_j'p_{\min}h_j^d\right)^2}\right)^{1/2}\notag\\
	&\leq
	\tilde{C}_2\max_{1\leq j\leq m}\frac{1}{m|D_j|h_j^d}\notag.
\end{align}
This completes the proof of Lemma \ref{Lemma:expectation-weight}. $\Box$

Based on the above three lemmas, we are in a position to present the following proposition.

\begin{proposition}\label{Proposition:important11111}
	Let $k\in\mathbb N$,   $x\in\mathbb I^d$ and   $\tilde{f}_{D,\tau}(x)$ be defined by \eqref{rewrite-f}     with $W_{j,h_j,x_{i,j}}(x)$ being given in Table \ref{Tab:application1} and $\tau\geq h^{2r}_j/|D|$ for any $j=1,\dots,m$.  If Assumption \ref{Assumption:distribution} holds,  then for any  $x$ satisfying that there exists at least a $j$ such that  $|f_{D_j,h_j}(x)|\geq \tau\geq h_j^{2r}/{|D|}$, there holds 
	\begin{equation}\label{proposition:generalization error}
	E[(\tilde{f}_{D,\tau}(x)-f^\diamond(x))^2]\leq  \tilde{C}_3(\tilde{c}_jh_j)^{2r}+ |D|^{-1}+   \max_{1\leq j\leq m}\frac{1}{m|D_j|h_j^d}
	\end{equation}
	where $\tilde{C}_3$ is a constant depending only on $d,r,p_{\min},p_{\max},c_0$ and $\|f^\diamond\|_{L^\infty}$.
\end{proposition}

{\bf Proof.}
For $\tau\geq h_j^{2r}/{|D|}$, according to (\ref{rewrite-f}), we have
$$
\tilde{f}_{D,\tau}(x)=\sum_{j=1}^m\sum_{i=1}^{|D_j|}\mathcal W_{\tau,D_j, {h}_j,x_{i,j}}(x)y_{i,j}.
$$
Set 
\begin{equation}\label{noiseless-version}
\tilde{f}^*_{D,\tau}(x)= \sum_{j=1}^m\sum_{i=1}^{|D_j|}\mathcal W_{\tau,D_j, {h}_j,x_{i,j}}(x)f^\diamond(x_{i,j}).
\end{equation}
Then, we have
\begin{align}\label{error-decomposition-3}
&&(\tilde{f}_{D,\tau}(x)-f^\diamond(x))^2\leq 2(\tilde{f}^*_{D,\tau}(x)-f^\diamond(x))^2
+ 
2(\tilde{f}_{D,\tau}(x)-\tilde{f}^*_{D,\tau}(x))^2.
\end{align}
Since there exists a $j$ such that $|f_{D_j,h_j}(x)|\geq \tau$, we have from   (\ref{weight:summation-big}) that 
$$
\tilde{f}^*_{D,\tau}(x)-  f^\diamond(x)
=
\sum_{j=1}^m\sum_{i=1}^{|D_j|}\mathcal W_{\tau,D_j, {h}_j,x_{i,j}}(x)(f^\diamond(x_{i,j})-  f^\diamond(x)).
$$
Hence
\begin{align}\label{p.sample-error-3}
&E[(\tilde{f}^*_{D,\tau}(x)-  f^\diamond(x))^2]
\leq
2E\left[
\left(\sum_{i,j,x_{i,j}\in B_{\tilde{c}_jh_j}(x)} \mathcal
W_{\tau,D_j,h_j,x_{i,j}}(x)(f^\diamond(x)-f^\diamond(x_{i,j}))\right)^2\right] \nonumber\\
&+
2E\left[
\left(\sum_{i,j,x_{i,j}\notin B_{\tilde{c}_jh_j}(x)}\mathcal W_{\tau,D_j,h_j,x_{i,j}}(x)(f^\diamond(x)-f^\diamond(x_{i,j}))\right)^2\right]\nonumber\\
&=:
2(\mathcal S_{1}+\mathcal S_{2}).
\end{align}
Due to \eqref{lip}, we get from (\ref{weight:summation-big})  that
\begin{equation}\label{p.C-1}
\mathcal S_{1}\leq c_0^2(\tilde{c}_j h_j)^{2r}.
\end{equation}
It follows from the H\"{o}lder inequality that
\begin{align}
	&
	\left(\sum_{i,j,x_{i,j}\notin\cap B_{\tilde{c}_jh_j}(x)}\mathcal W_{\tau,D_j,h_j,x_{i,j}}(x )(f_\rho(x )-f^\diamond(x_{i,j}))\right)^2\notag\\
	&\leq
	\left(\sum_{i,j,x_{i,j}\notin B_{\tilde{c}_jh_j}(x)}\mathcal W_{\tau,D_j,h_j,x_{i,j}}(x)\right)
	\left(\sum_{i,j,x_{i,j}\notin B_{\tilde{c}_jh_j}(x) }\mathcal W_{\tau,D_j,h_j,x_{i,j}}(x)(f^\diamond(x)-f^\diamond(x_{i,j}))^2\right)\notag.
\end{align}
Then it follows from (\ref{weight:summation-big}), (\ref{weight:localization-big}) and $\xi_j<(|D||D_j|M)^{-1}$  that
\begin{align}\label{p.C-2}
&\mathcal S_{2}
\leq
4\|f^\diamond\|_{L^\infty}^2\xi_jE\left[
\sum_{j=1}^m\sum_{i=1}^{|D_j|}\mathcal W_{\tau,D_j, {h}_j,x_{i,j}}(x)I_{\|x -x_{i,j}\|\geq \tilde{c}_jh_j}
\right]\nonumber\\
&\leq
4\|f^\diamond\|_{L^\infty}^2\max_{1\leq j\leq m}|D_j|\xi_j
\leq 4\|f^\diamond\|_{L^\infty}^2M^{-1}|D|^{-1}.
\end{align}
Plugging (\ref{p.C-1}) and (\ref{p.C-2}) into (\ref{p.sample-error-3}), we have
\begin{equation}\label{approximation-error-3}
E[(\tilde{f}^*_{D,\tau}(x)-  f^\diamond(x))^2]
\leq
2c_0^2(\tilde{c}_jh_j)^{2r}+8\|f^\diamond\|_{L^\infty}^2M^{-1}|D|^{-1}.
\end{equation}  
Noting further that $|f_{D_j,h_j}(x)|\geq \tau$ with $\tau\geq h_j^{2r}/{|D|}$ implies $B_{\tilde{c}_jh_j}(x)\cap \Lambda_j\neq\varnothing$,
we obtain from  (\ref{weight:summation-big}) that
$$
(\tilde{f}_{D,\tau}(x)-\tilde{f}^*_{D,\tau}(x))^2
=
\left(\sum_{j=1}^m\sum_{i=1}^{|D_j|}\mathcal W_{\tau,D_j,h_j,x_{i,j}}(x)(y_{i,j}-f_\rho(x_{i,j}))\right)^2.
$$
It thus follows from $f^\diamond(x_{i,j})= E[y_{i,j}|x_{i,j}]$ that
\begin{align}
	&E[(\tilde{f}_{D,\tau}(x )-\tilde{f}^*_{D,\tau}(x ))^2]
	\leq
	4\|f^\diamond\|_{L^\infty}^2 E\left[\sum_{j=1}^m\sum_{i=1}^{|D_j|}\mathcal W^2_{\tau,D_j,h_j,x_{i,j}}(x)\right]\notag.
\end{align}
Then, Lemma \ref{Lemma:expectation-weight} implies
\begin{align}\label{sample-error-2222}
E[(\tilde{f}_{D,\tau}(x)-\tilde{f}^*_{D,\tau}(x))^2]
\leq
4\|f^\diamond\|_{L^\infty}^2\tilde{C}_2\max_{1\leq j\leq m}\frac{1}{m|D_j|h_j^d}.
\end{align}
Plugging \eqref{sample-error-2222} and \eqref{approximation-error-3} into \eqref{error-decomposition-3}, we get 
$$
E[(\tilde{f}_{D,\tau}(x)-f^\diamond(x))^2]
\leq 4c_0^2(\tilde{c}_jh_j)^{2r}+16\|f^\diamond\|_{L^\infty}^2M^{-1}|D|^{-1}+ 8\|f^\diamond\|_{L^\infty}^2\tilde{C}_2\max_{1\leq j\leq m}\frac{1}{m|D_j|h_j^d}.
$$
This completes the proof of Proposition \ref{Proposition:important11111} with $\tilde{C}_3:=4\max\{c_0^2,4\|f^\diamond\|_{L^\infty}^2M^{-1},2\|f^\diamond\|_{L^\infty}^2\tilde{C}_2\}$. $\Box$

{\bf Proof of Proposition \ref{prop:TQMA-model}.}
Let $v$ be a random variable that follows the uniform distribution on the interval  $[a, b]$ with $a<b$. 
Under TQMA, the tree depth $k$ divides $[a, b]$ into $2^{k}$ sub-intervals. $v$ is then anonymized by the nearest midpoint  of these  sub-intervals, denoted as $v^{TQMA(k)}$. We then have $0\leq \|v - v^{TQMA(k)}\|_2 \leq \frac{b-a}{2^{k+1}}$.
Then, for 
$2\mu\in[0, (b-a)2^{-(k+1)}]$,
there holds
\begin{equation}
P(\|v - v^{TQMA(k)}\|_2 \leq 2\mu) \leq \frac{2\mu}{\frac{b-a}{2^{k+1}}} = \frac{\mu 2^{k+2}}{b-a}.
\end{equation}
For $2\mu > (b-a)2^{-(k+1)}$, there holds $P(\|v - v^{TQMA(k)}\|_2 \leq 2\mu)=1$.

{\bf Proof of Proposition \ref{prop:BSTD-model}.}
For any $j=1,\dots,m$ and $\ell=1,\dots,|D_j|$, it follows from the definition of BSTD in Algorithm 2 that 
$$
P[f_{D_j}^{p_\mathrm{lower},p_\mathrm{upper}}(x_\ell^{\mbox{fake}})=f_{D_j}(x_\ell^{\mbox{fake}})]\leq\frac1{p_\mathrm{upper}-p_\mathrm{lower}+1},
$$
implying \eqref{probability-bound} directly.  This completes the proof of Proposition \ref{prop:BSTD-model}. $\Box$

We then use Proposition \ref{Proposition:important11111} to prove  Theorem  \ref{Theorem:Optimal} as follows.

{\bf Proof of Theorem \ref{Theorem:Optimal}.}  
Due to \eqref{synthesis_form_active}, we have
$\tilde{f}_{D}=\tilde{f}_{D,1/{|D|}}$ with $\hat{h}_j=h_j^{\log_{|D_j|}|D|}$.
It follows from Proposition \ref{Proposition:important11111} with  $\tau=\frac1{|D|}$ and $x=x^{TQMA(k)}$ that if     $x^{TQMA(k)}\in\mathbb I^d$ satisfying that there exists at least a $j$ such that  $|f_{D_j,h_j}(x^{TQMA(k)})|\geq|D|^{-1}$ then
$$
E[(\tilde{f}_{D}(x^{TQMA(k)})-f^\diamond(x^{TQMA(k)}))^2]\leq  \tilde{C}_3(\tilde{c}_j\hat{h}_j)^{2r}+ |D|^{-1}+   \max_{1\leq j\leq m}\frac{1}{m|D_j|\hat{h}_j^d}.
$$
Noting $\tilde{c}_j\geq 1$, $|D_1|\sim\dots\sim |D_m|$ and $h_j\sim |D_j|^{-1/(2r+d)}$, we obtain from the above estimate that 
\begin{equation}\label{1.first-term}
E[(\tilde{f}_{D}(x^{TQMA(k)})-f^\diamond(x^{TQMA(k)}))^2]\leq  3\tilde{C}_3 \tilde{c}_j^{2r}|D|^{-\frac{2r}{2d}}.
\end{equation}
But Assumption \ref{Assumption:distribution} and the definition of TQMA yield
\begin{equation}\label{1.second-term}
(f^\diamond(x^{TQMA(k)})-f^\diamond(x))^2\leq c_0^2\|x^{TQMA(k)}-x\|^{2r}\leq c_0^22^{-2r(k+1)}.
\end{equation}
Hence, we obtain from \eqref{1.first-term} and \eqref{1.second-term} that 
\begin{align}
	& E[(\tilde{f}_{D}(x^{TQMA(k)})-f^\diamond(x))^2]
	\leq 
	2 E[(\tilde{f}_{D}(x^{TQMA(k)})-f^\diamond(x^{TQMA(k)}))^2]
	+
	2(f^\diamond(x^{TQMA(k)})-f^\diamond(x))^2\notag\\
	&\leq
	6\tilde{C}_3 \tilde{c}_j^{2r}|D|^{-\frac{2r}{2d}}
	+2c_0^22^{-2r(k+1)}\notag.
\end{align}
Note that the  global estimator $\tilde{f}_{D}(x)$ when using BSTD mechanism is almost the same as that  without using BSTD. The only difference is that in the qualification step, when not using BSTD, CMPP 
uses $|{f}_{D_j, \hat{h}_j}(x)|\geq \frac1{|D|}$ as the active rule, while when using BSTD, CMPP 
uses $|\frac{|D_j|}{|D|}{f}_{D_j, \hat{h}_j}(x)| \geq \frac{|D_{j}|}{|D|^2}$ as the active rule.
Therefore, for   $k\geq \frac{1}{4r+2d}\log_2|D|-1$, we have  
\begin{equation}\label{1.third-term}
E[(\tilde{f}_{D}(x^{TQMA(k)})-f_\rho(x))^2]\leq 
\tilde{C}_4\max_{1\leq j\leq m}\tilde{c}_{j}^{2r}|D|^{-\frac{2r}{2r+d}},
\end{equation}
where $\tilde{C}_4:=6\tilde{C}_3+2c_0^2.$ 
%
Then we obtain
\begin{equation}\label{thm_TQMA_log}
C_1 |D|^{-\frac{2r}{2r+d}}  
	\leq \mathcal U_{\mathcal M^{r,c_0}_{p_{\min},p_{\max}}}(\tilde{f}_{D},x^{TQMA(k)})\leq   C_2|D|^{-\frac{2r}{2r+d}} \log^{2r}|D|.
\end{equation}
TQMA with tree depth $k$ divides $[a, b]$ into $2^{k}$ sub-intervals. Each $x_i$ then takes the center point of its corresponding sub-interval, denoted as $x_i^{TQMA(k)}$, as its anonymous value. 
According to Proposition \ref{prop:TQMA-model} and Assumption \ref{Assumption:distribution} that
$$
  P(\|x_i- x_i^{TQMA(k)} \|_2) \leq \frac{ 2^{k+2}\mu p_{\max} }{b-a},
$$ we then have
%
\begin{align}
	&CO(\Xi_N,\Xi_N^{TQMA(k)}, \mu)
	=
	\frac{\sum_{i=1}^{N}I_{\|x_i- x_i^{TQMA(k)} \|_2\leq2\mu}}{N} \times 100\%
	=
	\frac{\sum_{i=1}^{N}P(\|x_i- x_i^{TQMA(k)} \|_2 \leq 2\mu)}{N} \times 100\% \notag\\
	&=
	P(\|x_i- x_i^{TQMA(k)} \|_2 \leq 2\mu) \times 100\%
	\leq
	\frac{  2^{k+2}\mu p_{\max} }{b-a} \times 100\%.
\end{align}

Due to the definition of BSTD, it follows that except for at most $p_\mathrm{lower}-1$ doctors, all doctors have changed their submitted predictions. Since $p_\mathrm{lower}\geq 2$, we have all these doctors cannot be linked. We can derive 
    \begin{equation}\label{RL_bound}
RL\Big(\{f_{B_j}(x_i^{TQMA(k)})\}_{j=1}^m,\ \{f_{B_j}^{p_{\text{lower}}, p_{\text{upper}}}(x_i^{TQMA(k)})\}_{j=1}^m\Big)\leq  \frac{100(p_\mathrm{lower}-1)}{m}\%.
 \end{equation}
The remaining thing is to prove the bound of $\mathcal U_{\mathcal M^{r,c_0}_{p_{\min},p_{\max}}}(\tilde{f}_D^{\mkern2mu p_\mathrm{lower},p_\mathrm{upper}},x^{\text{TQMA}(k)}))$.
This completes the proof of Theorem \ref{Theorem:Optimal}. We remove the details for the sake of brevity. $\Box$

\end{APPENDIX}

\end{document}